\documentclass{article}

\usepackage{arxiv}

\usepackage[utf8]{inputenc} % allow utf-8 input
\usepackage[T1]{fontenc}    % use 8-bit T1 fonts
\usepackage{hyperref}       % hyperlinks
\usepackage{url}            % simple URL typesetting
\usepackage{booktabs}       % professional-quality tables
\usepackage{amsmath,amssymb,amsfonts}
\usepackage{mathtools}
\usepackage{algorithmic}
\usepackage[ruled,vlined]{algorithm2e}
\usepackage{textcomp}
\usepackage{comment}
\usepackage{booktabs}
\usepackage{caption}
\usepackage{subcaption}
\usepackage{nicefrac}       % compact symbols for 1/2, etc.
\usepackage{microtype}      % microtypography
\usepackage{lipsum}
\usepackage{graphicx}
\usepackage{cite}
\title{A dynamic interactive learning framework for automated 3D medical image segmentation}

\author{
 Mu Tian \\
  School of Biomedical Engineering\\
  Medical School\\
  Shenzhen University\\
  Shenzhen, China 518000 \\
  \texttt{mtian@szu.edu.cn} \\
   \And
Xiaohui Chen \\
  School of Biomedical Engineering\\
Medical School\\
Shenzhen University\\
Shenzhen, China 518000 \\
\texttt{2060241021@email.szu.edu.cn} \\
  \And
Yi Gao \\
  School of Biomedical Engineering\\
  Medical School\\
  Shenzhen University\\
  Shenzhen, China 518000 \\
  \texttt{gaoyi@szu.edu.cn} \\
}

\begin{document}
\maketitle
\begin{abstract}
Many deep learning based automated medical image segmentation systems, in reality, face difficulties in deployment due to the cost of massive data annotation and high latency in model iteration. We propose a dynamic interactive learning framework that addresses these challenges by integrating interactive segmentation into end-to-end weak supervised learning with streaming tasks. We develop novel replay and label smoothing schemes that overcome catastrophic forgetting and improve online learning robustness. For each image, our multi-round interactive segmentation module simultaneously optimizes both front-end predictions and deep learning segmenter. In each round, a 3D “proxy mask” is propagated from sparse user inputs based on image registration, serving as weak supervision that enable knowledge distillation from the unknown ground truth. In return, the trained segmenter explicitly guides next step’s user interventions according to a spatial residual map from consecutive front/back-end predictions. Evaluation on 3D segmentation tasks (NCI-ISBI2013 and BraTS2015) shows that our framework generates online learning performances that match offline training benchmark. In addition, with a $\sim$62\% reduction in total annotation efforts, our framework produces competitive dice scores comparing to online/offline learning which equipped with full ground truth. Furthermore, such a framework, with its flexibility and responsiveness, could be deployed behind hospital firewall that guarantees data security and easy maintenance.
\end{abstract}

% keywords can be removed

\section{Introduction}\label{sec:introduction}
Existing works on automated segmentation systems~\cite{ref_2} for medical images majorly fall into one of two categories: interactive segmentation or fully automated segmentation.

Interactive segmentation incorporates user interventions to iteratively update the segmentation results. There are two fundamental desired properties in this type of methods: (1) the intuitiveness and convenience of user intervention to the software interface, and (2) the efficiency and accuracy of the underlying updating algorithm. Common types of user interventions include a set of clicks/stripes on the image that internally serve as ``seeds'' in outlining the target boundary. These seeds will then guide the algorithm to extract the target in multiple rounds. Conventional interactive segmentation methods were generally based on either active contour~\cite{ref_4, ref_6, ref_7, ref_8, ref_9, ref_10, ref_11, ref_12} or graph theory~\cite{ref_5, ref_13, ref_14, ref_15, ref_16, ref_17, ref_18, ref_19, ref_20, ref_21}. They are popular in clinical applications since they are fast responsive and light weighted. Well designed software tools have been deployed inside the hospital firewall and facilitate medical research without data leaving their origin. Such data-safety feature is critical to fulfill the current technological and legislative requirement. The downside is that they treat each new task (an image to be segmented) as an isolated case while ignoring correlations from similar tasks.

Contrastingly, fully automated segmentation methods construct predictive models that generate segmentations directly from raw images without external signals. In recent years deep learning based approaches have been dominating and many have achieved state-of-the-arts results. By exploiting local structures and spatial correlations, well designed networks successfully addressed the challenges in medical images including blurring, noise, low contrast and high variations~\cite{ref_2}. These networks in general should be trained sufficiently on adequate training data with high quality annotations. Constructing a whole deep learning development environment, training on specific target of interest, and deploying the trained model onto in-house datasets, are still formidable tasks for clinical researchers. As a result, clinical researchers are still taking the old route: sending the data out of the hospital to their engineering collaborators, teaching them what to look for, and waiting for the outcome. In this long turn-over process, even a slight modification of the target contouring will result in going through the entire process over again.

\subsection{Related Works}
\label{sec:related_works}

\subsubsection{Deep learning for Segmentation}

Various deep network architectures and learning strategies have been proposed for image segmentation. The most popular architectures include FCN~\cite{ref_26}, U-Net~\cite{ref_27} and its 3D extension~\cite{ref_28}, V-Net~\cite{ref_29}, DeepLab~\cite{ref_30}, and HighRes3DNet~\cite{ref_31}, etc. Fully convolutional operators, skip/residual connections and the fusion of low-high level information are key reasons for their successful performances. Attention mechanism proved helpful to control feature importance at different spatial/channel locations, and to further improve segmentation performances~\cite{ref_32}. For 3D cases, patch-based training is normally adopted to deal with data scarcity and balance memory consumption. Recently, adaptive patch selection, in addition to attention mechanism, were also found helpful to focus on more important regions and speed up convergence, which is helpful for extracting small organs~\cite{ref_33}. Carefully designed loss functions~\cite{ref_26, ref_29, ref_34, ref_35, ref_36} and weakly-supervised learning~\cite{ref_37, ref_38, ref_39, ref_40, ref_41} were also useful strategies for the case of limited data and class imbalance. 

\subsubsection{Interactive Segmentation}
\label{sec:related_works:inter}

Substantial number of conventional interactive segmentation methods were proposed, including those based on active contour~\cite{ref_4, ref_6, ref_7, ref_8, ref_9, ref_10, ref_11, ref_12} and graph algorithms~\cite{ref_5, ref_13, ref_14, ref_15, ref_16, ref_17, ref_18, ref_19, ref_20, ref_21}. They focus on making better use of local image features around user provided seeds, improving iteration speed of back-end algorithms, and handling multi-object cases. For example,~\cite{ref_4} proposed to adaptively learn image features around the seeds based on robust statistics to drive multiple active contours to evolve simultaneously, which utilized local features and addressed the issues in multi-target segmentation. Also,~\cite{ref_5} proposed fast grow cut with an adaptive Dijkstra algorithm that significantly reduced computational complexity in graph based algorithms. Both works~\cite{ref_4, ref_5} are delivered through an open-sourced integrated software environment, enabling their usage within the hospital firewall without data leaving their origin.

More recent works attempted to combine deep learning and interactive segmentation. Authors in~\cite{ref_42} proposed to transform user interactions to distance maps as extra channels to FCN and use graph cut to refine segmentation boundaries. It shows advantage in saving user workload in natural images. Similarly, DeepIGeoS~\cite{ref_22} also takes distance maps of user provided scribbles as extra CNN channels. They also adopt CRF as spatial regularizer to improve segmentation results. Moreover, a reinforcement learning based framework was proposed in~\cite{ref_23} that incorporate user interactions into trainable dynamic process. In~\cite{ref_42, ref_22, ref_23}, their training stages all require transformed user interactions as input, it is unclear how to generalized on unseen cases without user inputs. Also, they require ground truth labels to be present during some or all stages of training, and thus could hardly deal with situations of cold-start and streaming data.

Contrastingly, BIFSeg~\cite{ref_3} adopt user provided scribbles as label adjustment for image specific fine-tuning, that makes CNN adaptive to a particular unseen image with the possible aid of user provided scribbles. One advantage of BIFSeg is that it could generalize to unseen image classes with small number of user interventions. However, the user interaction step is isolated from training dynamics and therefore it has no difference to non-interactive deep learning should there be no user inputs during testing. Also, the fine-tuning step can only be effective after sufficient training on fully labeled images. Other methods such as 3D U-Net~\cite{ref_28} and DeepCut attempted to minimize user inputs by treating them as sparse annotations, yet they are not responsive to user inputs during test and still requires sufficient annotations before achieving reasonable performances. 

\subsubsection{Online Learning}

Online continual learning~(OCL) is a popular family of strategies in dealing with streaming data. In deep learning era, training effective models online is more challenging due to the requirement of sufficient data to accommodate model complexity. In online setting, training samples arriving sequentially and each sample is normally seen once by the model. Catastrophic forgetting~\cite{ref_45, ref_46} is the key challenge for online learning since the distributions of labels and data itself could shift over time and the model is more affected by most recent instances. Therefore, during online training, resampling from past examples properly could help to balance information learned from past and new samples. Popular methods such as iCaRL~\cite{ref_47} , GDumb~\cite{ref_48} and MIR~\cite{ref_49} keep a buffer to store most important samples for the model, and they differ by enqueue/dequeue operations. There were also a few works studying OCL for segmentation~\cite{ref_50, ref_51, ref_52, ref_53}, and they all focused on the case of shifted classes/domains instead of handling streaming data with identical task. In our study, we care more about the image distribution shift along the same task/domain. To the best of our knowledge, this paper is the first to develop, discuss and integrate OCL dynamics with interactive 3D medical image segmentation.

\subsection{Our Contributions}
\label{sec:our_contributions}

We attempt to develop a novel framework. Such framework should integrate interactive segmentation and dynamic learning with the deep network based segmenter into a synergistic system that can be trained end-to-end. It should also be deployed into the end user, such as behind the hospital firewall, and facilitate the specific 3D image annotation on images acquired with possibly very special configurations. A list of desired properties should be considered in order for practical deployment.

% Realizing the time overhead and tediousness in collecting these manual annotations (especially in 3D), it is natural to merge the two streams of efforts and use deep networks as the backbone segmentor in an interactive framework [3, 22, 23]. Comparing to conventional interactive segmentation methods, the deep network module is better at capturing contextual features in the image through learning from user hints and a series of similar tasks, and thus could generalize better in unseen cases.

First, the system should generate sufficiently responsive, accurate and robust results with limited human resources. Due to the tediousness of drawing labels in 3D, the deployment of deep learning based systems would be a long process. Moreover, the annotation and model training may take place at different locations by different teams in separated steps, it's difficult to estimate how many volumes the physicians need to contour for a sufficient amount for model training. Furthermore, the distribution of the images and their labels could shift over time~\cite{ref_3}, due to possible changes in equipment, patient population and labeling protocols. Existing models trained in a batch wise fashion offline woule be difficult to adapt to the above changes in real time. 

Existing deep learning based interactive segmentation methods~\ref{sec:related_works:inter} attempted to solve one or more of the above challenges but there are still some gaps. First, it is unclear how to properly deal with cold start cases. Existing works did not show how they could start producing meaningful predictions instantly after a few rounds of user interactions without pre-training. Moreover, it is not straightforward how these algorithms could consistently generate segmentations on unseen images with zero user interactions: trained deep learning models could certainly produce some start-up segmentations on new images, but for systems~\cite{ref_42, ref_23, ref_22} requiring user hints as auxiliary inputs during interactive training, a new sample with no user input would incur an inconsistency leading to the ignorance of the knowledge from training stage. Furthermore, without knowing the ground truth, it is difficult for existing methods to suggest users how to provide annotations in the next round. We could expect faster convergence if the back-end algorithm could guide clinicians to spend efforts on only important regions. Finally, how to dynamically update the system with sequentially arriving data is still a challenge. 

%In practice, we'd expect labeled images to be collected gradually from clinicians. The system will become less responsive if we periodically train models with offline batches after acquiring adequate data. Existing interactive segmentation methods did not discuss how to robustly update the system in near-real time, with sequentially arriving data.

We propose a dynamic interactive learning framework that addresses these challenges by integrating interactive segmentation into end-to-end weak supervised learning with streaming tasks. We develop novel replay and label smoothing schemes that overcome catastrophic forgetting and improve online learning robustness. For each image, our multi-round interactive segmentation module simultaneously optimizes both front-end predictions and deep learning segmenter. In each round, a 3D “proxy mask” is propagated from sparse user inputs based on image registration, serving as weak supervision that enable knowledge distillation from the unknown ground truth. In return, the trained segmenter explicitly guides next step’s user interventions according to a spatial residual map from consecutive front/back-end predictions.

Our work has four fold contributions:

First, the proposed framework supports dynamic learning with streaming data and provides near-real time inference, without requiring the existence of full ground truth throughout the process. Experiment results show that we could achieve sufficient accuracy with substantially reduced user labor.

Second, we present the first attempt to thoroughly discuss and experiment on the mechanisms of dynamic training combined with interactive medical image segmentations. We discuss optimal strategies not just about user interactions as in previous works, but also about model guided annotation, training with sequential data, and learning without ground truth labels. 

Third, our framework is highly flexible. It can be trained end-to-end in a unified stage. It also supports inference with or without user interventions, and generalizes to using different network architectures.  

Fourth, in addition to the technical contributions above, logistically, such a interlinked annotation and dynamic learning system could be deployed into the hospital firewall that guarantees ultimate data security and prevents the potential hurdles in cross-functional collaborations.

%\IEEEpubidadjcol

%%%%%%%%%%%%

\section{Methodology}
\label{sec:method}
\begin{figure*}[htb]
	\centering
	\includegraphics[width=\textwidth]{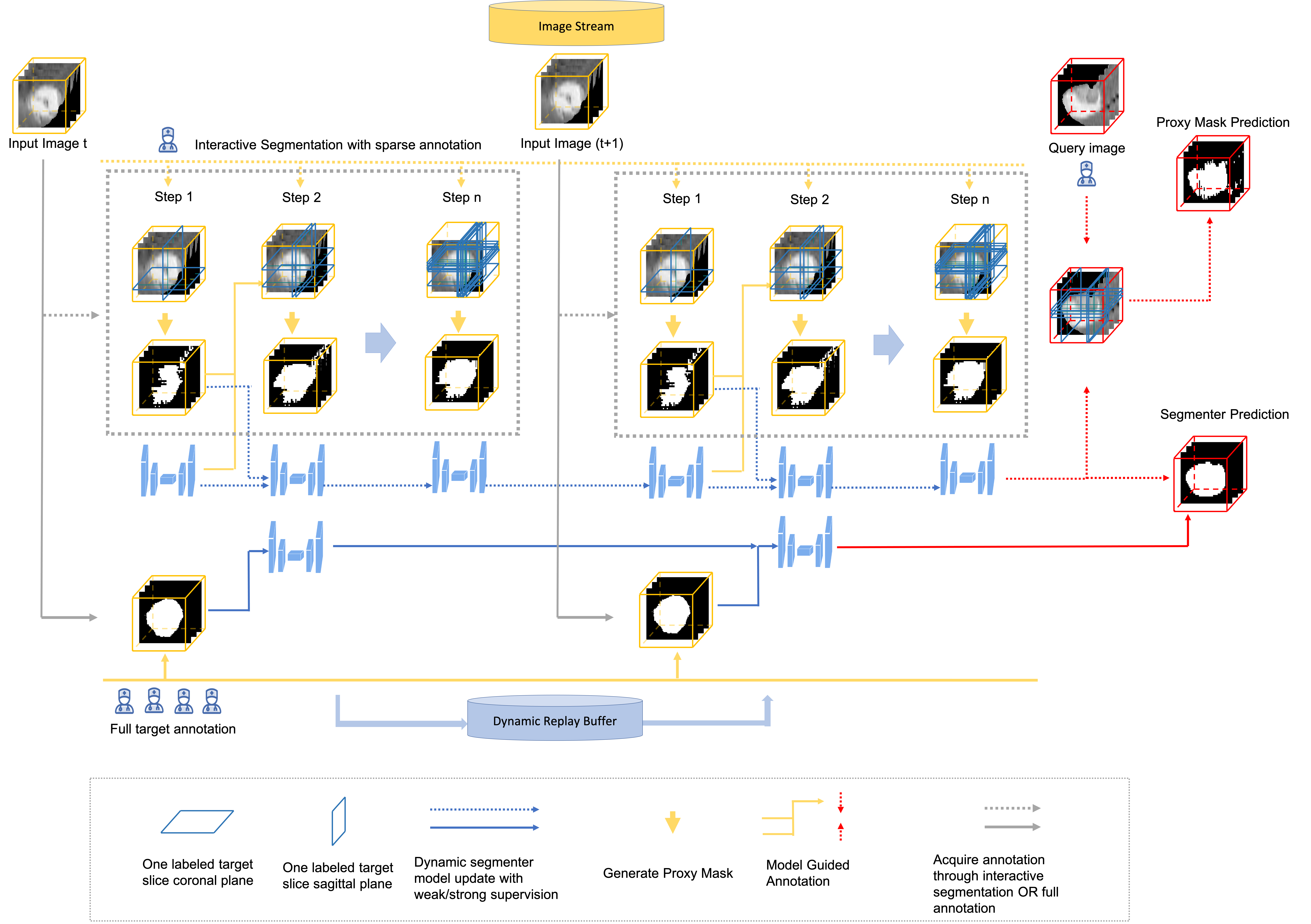} % don't need the ext-name
	\caption{A overview of our dynamic interactive learning framework. Detailed in sections~\ref{sec:method} and~\ref{sec:exp}}
	\label{fig:flowchart_dynaminter}
\end{figure*}
We illustrate the overview of our framework in Figure~\ref{fig:flowchart_dynaminter}. 3D volumes arrive sequentially from image stream and will either be labeled directly by physicians or enter interactive segmentation, based on practical needs. The dynamic learning module determines the mechanisms for training the underlying segmenter. To address catastrophic forgetting~\cite{ref_45, ref_46}, we propose a combined scheme including a novel replay mechanism, loss weighting and label smoothing. The interactive segmentation module allows users to provide sparse annotations iteratively. A label propagation technique is developed to transform user inputs into an approximate segmentation mask (a.k.a the proxy mask) of the full target, which is then used as weak supervision for dynamic learning. In contrast to the other deep learning based interactive segmentation methods such as~\cite{ref_42, ref_22, ref_23}, our setup could deal with cold start case and guarantee the consistency between training/testing stages, since the segmenter could start to incrementally produce meaningful predictions from the early interaction steps without knowing the ground truth. In addition, the model guided annotation component could provide explicit guidance to pinpoint spatial locations for next round of user interventions.

\subsection{Deep Learning Segmenter}

The focus of this work is not to design new network architectures. Our framework support any backbone segmentation networks. In this paper we built our network based on 3D V-Net (Figure 2 in~\cite{ref_29}) which is frequently adopted in interactive segmentation studies~\cite{ref_23}. Similar to U-Net~\cite{ref_27}, V-Net is another modern U-shaped fully convolutional architecture whose encoder-decoder design is capable of capturing low-to-high level information efficiently.

\subsection{Dynamic Learning}\label{sec:dynam_learn}

The major challenge of training deep networks dynamically is to balance the experience learned from past and current data. Most OCL works~\cite{ref_25} study online classification tasks. However, we found that typical OCL strategies successful for classification does not provide immediate gain for segmentation. For online segmentation, the variation in data distribution originates not just from the input images, but also from segmentation masks. Viewing as voxel level classification, we are dealing with sequential variation in much higher dimensional space than that in online classification. 

Let $T=\{(X_1, Y_1), (X_2, Y_2), \dots \}$ be the sequence of images $\{X_{i}\}$ defined on 3D domain $\Omega$ with binary ground truth masks $\{Y_i\}$ where $1$ being the foreground label and $0$ the background. Let $F(X) = f(X|\theta)$ be the automated segmenter that generate a probability map $P\in[0, 1]$ indicating the likelihood of each voxel belonging to the target. In supervised learning, we obtain $\theta$ (thus $F$) by minimizing some loss function $l(\theta|Y, F(X))$. Commonly used loss functions include binary cross entropy loss and dice loss. Let $T_t=\{(X_t, Y_t), i = 1, \dots, t\}$ and $F_t(X) = f(X|\theta_{T_t})$ be the model trained after seeing first $t$ samples in the sequence. Fundamentally, we need a mechanism to update $F_t$ to $F_{t+1}$. 

%\begin{align}
%	l_{\text{bce}}(\theta|Y, P) &= -\frac{1}{n}\sum[y\cdot \log{p} + (1 - y)\cdot \log{(1 - p)}] \\
%	l_{\text{dice}}(\theta|Y, P) &= 1 - \frac{2\sum(y\cdot p) + \epsilon}{\sum(y + p) + \epsilon} 
%	\label{eq:bce_and_dice_loss}
%\end{align}
%where $n$ is the total number of voxels and the sums are over the entire image domain. 

There are two naive ways for the update: (1) minimize over the entire data set so far $L(\theta | T_{t+1}) = \frac{1}{t+1}\sum_{i=1}^{t+1}l(\theta|Y_i, F(X_i))$. (2) minimize over just the new sample $L(\theta | \{(X_{t+1}, Y_{t+1})\}) = l(\theta | Y_{t+1}, F(X_{t+1}))$ and ignore the past. The first approach can not scale up with more samples arriving and the second approach suffers from catastrophic forgetting~\cite{ref_45, ref_46}. Different from offline training, the update should take constant cost with at most $O(1)$ revisits to the past sequence $T_{t}$.

In OCL, replay techniques~\cite{ref_54} maintain a fixed sized memory buffer $B_t, |B_t| \equiv b > 0$ storing part of past samples, the model is trained with the new sample and selected past samples (from buffer) together. The loss function becomes equation~\ref{eq:loss1} below:
\begin{align}
	L(\theta | \{(X_{t+1}, Y_{t+1})\}  \cup S_{t}) = \frac{1}{m+1}[\sum_{i=1}^{m}l(\theta|Y_{m_i}, F(X_{m_i})) + l(\theta|Y_{t+1}, F(X_{t+1}))]
	\label{eq:loss1}
\end{align}
where $S_t = \{(X_{m_i}, Y_{m_i})\}|_{i=1}^{m} \subset B_t$ is set of the $m \leq b$ selected past samples from buffer, and $m_i$'s are corresponding index mappings. In addition, the buffer $B_t$ should be updated to $B_{t+1}$ with the new sample. Popular strategies for sample selection and buffer update include experience replay~\cite{ref_54}, Maximally Interfered Retrieval (MIR)~\cite{ref_24}, GSS~\cite{ref_55} and GDumb~\cite{ref_48}. Though they showed effectiveness for classification, we observed sub-optimal performances when simply applying them for 3D segmentation. We propose new strategies for buffer update and sample selection, combined with adaptive weighting on loss functions and targets.

Let $B$ be the memory buffer at step $t$ with size $b>0$, algorithm~\ref{algo:buffer_update} is used to maintain $B$ such that we have a slowly decaying possibility of revisiting older samples and relatively higher chances for newer samples.
\begin{algorithm}
	\caption{Buffer Update at step $t$}
	\begin{algorithmic}[1]
		\STATE{\textbf{Input}: buffer $B(b)$, auxiliary buffer $B^{\star}(b^{\star})$ ($b^{\star} > b$), new sample $(X_{t}, Y_{t})$}
		\STATE{Set $j = \text{randint}(0, b^{\star})$}
		\IF{$j < b^{\star}$}
		\STATE{Set $B^{\star}[j] = (X_{t}, Y_{t})$}
		\ENDIF
		\STATE{Set $B(b) = \text{random.select}(B^{\star}(b^{\star}), b)$}
	\end{algorithmic}
	\label{algo:buffer_update}
\end{algorithm}

Note that this algorithm~\ref{algo:buffer_update} is similar, but different to both reservoir sampling~\cite{ref_56} and Gdumb~\cite{ref_48}. Each step, we retrieve $m<b$ samples from the buffer $B(b)$ to form $S_t$, then combine with the new sample $(X_{t+1}, Y_{t+1})$ for training. We use random sampling during retrieval and we found none of model assisted methods (MIR~\cite{ref_24}, GSS~\cite{ref_55}, etc) provided additional gains. 

Moreover, we introduce weights for loss function to control the importance of selected past/new samples. Also, we allow ``label smoothing'' on the target to encourage smooth adaptation of model predictions to data distribution drift. This is formulated in equation~\ref{eq:our_dynamic_loss} below:

\begin{align}
	L(\theta | \{(X_{t+1}, Y_{t+1})\} \cup S_{t}) = \frac{1 - \lambda_L(t)}{m+1}\sum_{i=1}^{m}l(\theta|\tilde{Y}_{m_i}, F(X_{m_i})) + \lambda_L(t) l(\theta|\tilde{Y}_{t+1}, F(X_{t+1}))
	\label{eq:our_dynamic_loss}
\end{align}

Here $\lambda_L(t) \in [0, 1]$ is the loss weight that balances the importance of residuals on past and new samples; $\hat{Y}(t) = (1 - \lambda_Y(t))F_t(X) + \lambda_Y(t)Y$ is the weighted target where $\lambda_Y(t)$ is the weight to control label smoothness, and $F_t(X)$ is the prediction on the current sample from the previous step's segmenter.

\subsection{User Interactions}\label{sec:user_inter}

Typical types of user inputs in existing interactive segmentation software include drawing clicks/stripes on 2D planes of the 3D volume. it normally takes multiple rounds of user intervention and system feedback to reach a satisfying result. It is thus important to optimize both the intervention and feedback loops to reduce overall user labor. This work does not intend to compare different forms of front-end user inputs, rather, we focus on the optimization problem that simultaneously improve user experience and the performance of back-end algorithms. Some work~\cite{ref_3} provided the actual user time collected from real physicians. But it may be difficult to control variations in the working process considering software design and user habits. Therefore, we use one fixed form of user inputs across our experiments and compute user ``cost'' in a relative manner.

We assume that user draw 2D boundaries of the target slice by slice to eventually form the 3D mask. Suppose the target volume has the dimension $(n_x, n_y, n_z)$ along axial, sagittal and coronal planes, the drawing could be done along any one of the three directions. It is reasonable to assume a constant cost, defined in equation~\ref{eq:labor_3d} of labeling the entire target regardless of which plane to start with:
\begin{align}
	C(Y) = n_xC_x(Y) = n_yC_y(Y) = n_zC_z(Y)
	\label{eq:labor_3d}
\end{align}	
where $C_{x/y/z}(Y)$ is the constant cost, for simplicity, of drawing the target boundary of one slice along plane $x/y/z$. 

In conventional settings, the training samples are fully labeled at once. We are interested in achieving similar level of performances with only partial annotation. Specifically, let $\Gamma_{x/y/z}, |\Gamma_{x/y/z}| < n_{x/y/z}$ be the indices of labeled target slices along plane $x/y/z$, the cost is then defined in equation~\ref{eq:labor_partial} below:
\begin{align}
	C_{\Gamma}(Y) = |\Gamma_x|C_x(Y) + |\Gamma_y|C_y(Y) + |\Gamma_z|C_z(Y) = (\frac{|\Gamma_x|}{n_x} + \frac{|\Gamma_y|}{n_y}  + \frac{|\Gamma_z|}{n_z})C(Y)
	\label{eq:labor_partial}
\end{align}	

Our interactive segmentation module aims at, embedded in dynamic learning setting, finding the best combinations of $\Gamma$ to maximally reduce user labor: i.e. $\text{argmin}_{\Gamma}C_{\Gamma}(Y)$ such that the resulting segmentation is satisfying. 

Two fundamental questions must be addressed: (1) how the system could learn from the partial annotation without knowing the ground truth and (2) how our system could help to automatically reduce the search space of $\Gamma$. We will discuss (1) and (2) in sections~\ref{sec:proxy_mask} and~\ref{sec:model_guidance} respectively.

\subsection{Learning with Proxy Masks}
\label{sec:proxy_mask}
The proposed partial annotation is similar to~\cite{ref_28}, which showed that a 3D U-Net could produce approximate 3D masks after training on many sparsely labeled samples with weighted loss. However, the conclusion in~\cite{ref_28} only works in offline training where the sufficient sparse annotations are instantly available. This is fundamentally different to our dynamic interactive setting where responsiveness is critical.

In our framework, we propagate each step's partial annotations to an approximate 3D mask (called ``proxy mask'') serving as weak supervision. We will show that, empirically, model trained with proxy masks interactively converges to unknown true labels rather than proxy masks. We use image registration as the basic operation to propagate annotated 2D slices onto unlabeled areas. Formally, given a ``fixed'' image $X^f$  and ``moving'' image $X^m$, we search for a transformation $T$ such that the transformed $X^m$ is closest to $X^f$ under some metric, as in equation~\ref{eq:image_reg}:
\begin{align}
	T^{*}_i = \text{argmin}_{T:\Omega \to \Omega} D(X^f, X^m \circ T)
	\label{eq:image_reg}
\end{align}
where $D$ is a dis-similarity metric between two images. $T$ can be solved by iterative procedures such as gradient descent or Newton method. In this work, we take $T$ from the family of affine/deformable transformations and use mutual information as the optimization objective. Algorithm~\ref{algo:slice_prop_register} generates a 3D proxy mask from partially labeled slices along one plane (assuming axial W.L.O.G., $\Gamma_x$). 

\begin{algorithm}
	\caption{Generate proxy mask along one plane}
	\begin{algorithmic}[1]
		\STATE{\textbf{Input}: image $X\in \mathbb{R}^{d_x\times d_y \times d_z}$, labeled target slices indices $\Gamma_x$ and their 2D masks $\{Y^{i}, i \in \Gamma_x\}$} 
		\STATE{\textbf{Output}: 3D Proxy mask $\Upsilon_{x} = \{\Upsilon_{x}^{j}, j = 1, \dots, d_x  \}$}
		\STATE{Set $\Upsilon_{x}=\textbf{NULL}$}
		\FOR{$j \in 1, \dots, d_x$}
		\IF{$j \in \Gamma_x$}
		\STATE{Set $\Upsilon_{x}^{j} = Y^{j}$}
		\ELSE
		\STATE{Set $k^{\star} = \text{argmin}_{k \in \Gamma_x} | j - k |$}
		\STATE{Solve $T^{\star} =  \text{argmin}_{T} D(X^j, X^{k^{\star}} \circ T)$}
		\STATE{Set $\Upsilon_{x}^{j} = Y^{k^{\star}} \circ T^{\star}$}
		\ENDIF
		\ENDFOR
	\end{algorithmic}
	\label{algo:slice_prop_register}
\end{algorithm}

In fact, searching for $k^{*}$ takes $O(d_x)$ as a pre-process step and therefore the algorithm takes $KO(d_x)$ where $K$ is the constant cost from image registration/transformation. In interactive mode, new labeled slices are enqueued into $\Gamma$ incrementally and we only need to update for $\{j = 1, \dots, d_x | \text{argmin}_{k \in \Gamma_x \cup \Gamma_x^{\prime}} | j - k | \notin \Gamma_x, j \notin \Gamma_x^{\prime}\}$ where $\Gamma_x^{\prime}$ is the new slices labeled in next round. Unlike linear complexity in static mode where $\Gamma_x$ is given all at once, the overall asymptotic cost for interactive mode is between $O(d_x)$ and $O(d_x^2)$ as $|\Gamma_x| \rightarrow d_x$. In average case, assuming $\Gamma_x$ is distributed uniformly, the approximate cost for the $t^{\text{th}}$ update is $\sim O(d_x / t)$ assuming $|\Gamma_x| = t - 1$ and $|\Gamma_x^{\prime}| = 1$. Note that the algorithm can be executed in parallel. 

Note that without knowing the real target, algorithm~\ref{algo:slice_prop_register} tends to produce excessive foreground slices outside of the true boundaries. It is beneficial to execute the algorithm along two or more planes and merge the generated proxy masks to neutralize the false positive areas. We found it sufficient to do along two planes, for example, using $\Upsilon = (\Upsilon_{x} + \Upsilon_{y}) / 2$ could effectively suppress the excessive false positives with cut-off threshold 0.5.

The proxy mask is used in two different ways. First, $\Upsilon$ itself can be returned to the front-end as approximate segmentations. Second, it is used as a weak label to train the segmenter $F$. In both cases, we seek simultaneous improvements of $\Upsilon$ and $F$. Generally we use $Dice(Y, \Upsilon)$ to measure how well $\Upsilon$ approximates the unknown $Y$. Though without theoretical proofs, we have an empirical conclusion in all experiments that $Dice(Y, \Upsilon)$ increases monotonically with $|\Gamma_{x/y}|$. Formally, we look for:

\begin{align}
	F^{\star} &= \text{argmax}_{F | \Gamma}Dice(Y, F) \ \ \text{s.t.} \ |\Gamma|  \leq \gamma \label{eq:inter_learning_opt_3} \\
	\Gamma^{\star} &= \text{argmax}_{\Gamma}Dice(Y, \Upsilon) \ \ \text{s.t.} \ |\Gamma|  \leq \gamma  \label{eq:inter_learning_opt_1} \\	
	\Gamma^{\star} &= \text{argmin}_{\Gamma}|\Gamma| \ \ \text{s.t.} \ Dice(Y, \Upsilon) > \rho  \label{eq:inter_learning_opt_2}
\end{align}

where $\gamma>1$ and $0<\rho<1$ represents maximum number of labeled slices allowed (equivalently, user ``quota'')  and the Dice score threshold. Equations~\ref{eq:inter_learning_opt_1} and~\ref{eq:inter_learning_opt_2} are suitable for slightly different situations in practice, which will be discussed in experiments. Unlike equation~\ref{eq:inter_learning_opt_2}, it is difficult to set a lower bound constraint for $Dice(Y, F)$ since it does not have a generally achievable upper bound.

Even assuming knowing $Y$, searching for above solutions takes combinatorial complexity. In practice, users cannot ``undo'' their actions so each round of inputs is counted into $\Gamma$. Therefore, a one-pass heuristic algorithm is needed. This section discusses equation~\ref{eq:inter_learning_opt_3} and section~\ref{sec:model_guidance} will discuss~\ref{eq:inter_learning_opt_1} and~\ref{eq:inter_learning_opt_2}. Algorithm~\ref{algo:interactive_learning} below formalizes a heuristic incremental procedure for equation~\ref{eq:inter_learning_opt_3}. 

\begin{algorithm}
	\caption{Interactive Learning}
	\begin{algorithmic}[1]
		\STATE{\textbf{Input}: Image $X$, initial segmenter $F$, quota for user inputs $\gamma$, round of interactions $n_{\text{inter}}$, buffer size $b$, sample selection size $s$}
		\STATE{\textbf{Output}: Proxy mask $\Upsilon$, updated segmenter $F$}
		\STATE{Set $\Gamma = \{ \}$, $i = 0$, $\Upsilon=\mathbf{0}$, buffer $B=\{ \}$, sample set $S=\{ \}$, user front-end: $FN=\{X, \Upsilon, \Gamma, F(X)\}$}
		\WHILE{$|\Gamma| < \gamma$ and $i < n_{\text{inter}}$}
		\STATE{\textbf{User}: Decide new slices to label $\Gamma^{\prime}$ according to front-end $FN$}
		\STATE{\textbf{User}: Draw labels for slices in $\Gamma^{\prime}$}
		\STATE{Set $\Gamma = \Gamma \cup \Gamma^{\prime}$}
		\STATE{Set $\Upsilon$ with $\Gamma^{\prime}$ (algorithm~\ref{algo:propose_slice_to_label})}	
		\STATE{Set $\mathbf{W}, \lambda_L, \lambda_Y, n_F(i)$}
		\FOR{$j = 1, \cdots, n_F(i)$}
		\STATE{Set $\theta = \theta - \eta \bigtriangledown_{\theta} L(\theta | \{(X, \tilde{\Upsilon})\} \cup S)$}
		\ENDFOR	
		\STATE{Set $B, S$ (algorithm~\ref{algo:buffer_update})}
		\STATE{Update $FN=(X, \Upsilon, \Gamma, F(X))$}
		\STATE{Set $i = i + 1$}
		\ENDWHILE
	\end{algorithmic}
	\label{algo:interactive_learning}
\end{algorithm}

Note that $Y$ is not available beyond the queries incurred by $\Gamma$. The algorithm involves users (real experts) to decide where and how to provide inputs according to system feedback ($FN$) in each round. After labeling, the proxy mask is generated for training the automated segmenter $F$. Though this algorithm discusses the case for a single image, multi-round interaction inherently leads to a dynamic learning task. Therefore, we adopt algorithm~\ref{algo:buffer_update} and equation~\ref{eq:our_dynamic_loss} with properly chosen parameters including sizes of $B, S$, and $\lambda_{L/Y}$ in the weighting scheme. In addition, it is expected that the confidence (likelihood of a correct prediction) in $\Upsilon$ is higher near $\Gamma$. Therefore, we introduce a spatial weight in the loss (used in algorithms~\ref{algo:interactive_learning} and~\ref{algo:dynam_inter_learn}) in equation~\ref{eq:spatial_weighted_loss}:

\begin{align}
	L(\theta | \{(X_{t+1}, Y_{t+1})\} \cup S_{t}) &= \frac{1 - \lambda_L(t)}{m+1} \langle{ \mathbf{W}(\tilde{Y}_{m_i} | \Gamma_{m_i}) ,  \mathbf{l}(\theta|\tilde{Y}_{m_i}, F(X_{m_i})) }\rangle  \nonumber \\
	&+ \lambda_L(t) \langle{ \mathbf{W}(\tilde{Y}_{t+1} | \Gamma_{t+1})  ,   \mathbf{l}(\theta|\tilde{Y}_{t+1}, F(X_{t+1}))  }\rangle
	\label{eq:spatial_weighted_loss}
\end{align}

where $\mathbf{W}$ represents the confidence map, $\mathbf{l}$ is the voxel-wise form of the loss and $\langle{,}\rangle$ is inner product. Note that we continue to use $Y$ to represent ``target'' W.L.O.G, including cases for strong (true $Y$, section~\ref{sec:dynam_learn}) and weak supervisions (algorithm~\ref{algo:interactive_learning}). Equations~\ref{eq:spatial_weight_forms_1},~\ref{eq:spatial_weight_forms_2} and~\ref{eq:spatial_weight_forms_3} list the generic form of $\mathbf{W}$ along with two commonly used instantiations in our work:  
\begin{align}
	\mathbf{W}(\mathbf{x}|\Gamma) &= g(d(\mathbf{x}, \Gamma))  \label{eq:spatial_weight_forms_1} \\
	\mathbf{W}(\mathbf{x}|\Gamma) &= e^{-\min_{\mathbf{y} \in \Gamma} \|\mathbf{x} - \mathbf{y}\| } \label{eq:spatial_weight_forms_2} \\
	\mathbf{W}(\mathbf{x}|\Gamma) &=  \omega \mathbf{1}_{\mathbf{x} \in \Gamma} + \mathbf{1}_{\mathbf{x} \in \Omega} \label{eq:spatial_weight_forms_3}
\end{align}
where $g$ is a non-increasing function and $d$ is a distance metric, $\omega>0$ is a parameter to control the confidences over $\Gamma$ and the other area. 

The choice of $\Gamma$ fundamentally determines, in part, the quality of $\Upsilon$ and thus $F$, as well as the overall turn-over time of the interactive process. In addition to labeling, users also spend time on choosing $\Gamma$. It is difficult to consistently quantify the cost of the decision process. In simulated studies, some work assumed identical user actions, for example, to label the key point of biggest residual connected component~\cite{ref_23}. However, this is based on a strong assumption that the ground truth is visible at user end throughout the process. In contrast, we do not impose this assumption, rather, we assumed random selection (the ``worst'' case~\cite{ref_23}) in default setting. The following section introduces a mechanism that free users from the decision process while improving overall performance and efficiency of interactive learning.

\subsection{Model guided annotations}
\label{sec:model_guidance}

\begin{figure*}[htb]
	\centering
	\includegraphics[width=\textwidth]{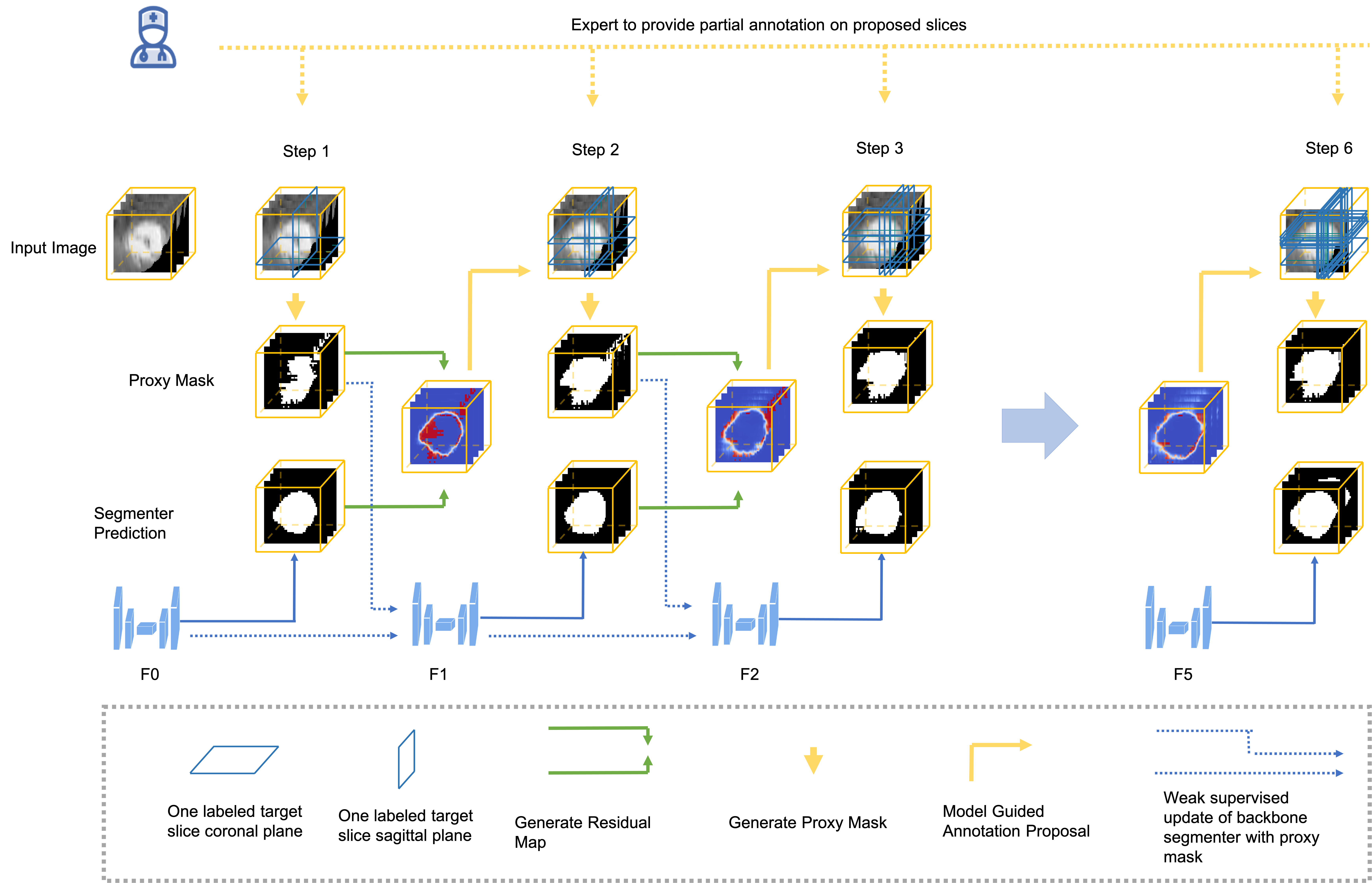} % don't need the ext-name
	\caption{A flowchart illustrating interactive segmentation with model guided annotations. The example input image has six rounds of interactive learning (Step 1 thru Step 6)}
	\label{fig:flowchart_inter}
\end{figure*}

We have shown how user interventions could assist training $F$, in return, we show how the segmenter could help users in the front-end. In fact, as an interactive process, the segmenter $F$ should naturally participate in the optimizations (~\ref{eq:inter_learning_opt_1}, ~\ref{eq:inter_learning_opt_2}): 

\begin{align}
	\Gamma^{\star} &= \text{argmax}_{\Gamma | F}Dice(Y, \Upsilon) \ \ \text{s.t.} \ |\Gamma|  \leq \gamma  \label{eq:inter_learning_opt_11} \\	
	\Gamma^{\star} &= \text{argmin}_{\Gamma | F}|\Gamma| \ \ \text{s.t.} \ Dice(Y, \Upsilon) > \rho  \label{eq:inter_learning_opt_21}
\end{align}

If we already have a ``perfect'' segmenter such that $F(X) \equiv Y$ we could formulate~\ref{eq:inter_learning_opt_11} in incrementally:

\begin{align}
	\max_{\Gamma^{\prime}}&(Dice(\Upsilon^{\prime},Y) - Dice(\Upsilon, Y))
	\label{eq:opt_search_slice_incremental}
\end{align}

In reality, we could solve the approximate problem:

\begin{align}
	\max_{\Gamma^{\prime}}&(Dice(\Upsilon^{\prime}, F(X)) - Dice(\Upsilon, F(X)))
	\label{eq:opt_search_slice_incremental_approx}
\end{align}

A ``better'' $F$ should provide a closer convergence to the original optimal solution. In fact, we have an empirical conclusion that $Dice(\Upsilon, F(X)) \leq Dice(Y, F(X))$. Therefore, solving~\ref{eq:opt_search_slice_incremental_approx} moves $F$ closer to the unknown $Y$ and thus incrementally approaches~\ref{eq:opt_search_slice_incremental}. Likewise, for such a combinatorial problem, algorithm~\ref{algo:propose_slice_to_label} provides a heuristic method.

\begin{algorithm}
	\caption{Model guided selection for $\Gamma_x^{\prime}$}
	\begin{algorithmic}[1]
		\STATE{\textbf{Input}: $F$, Current $\Gamma, \Upsilon$, number of new slices along each plane $\tau$}
		\STATE{\textbf{Output}: $\Gamma_x^{\prime}$}
		\STATE{Compute residual map $r(\cdot | \Upsilon, F, \Gamma)$}
		\STATE{Set $Pr = [], Pr_{\text{sum}}= 0, S = [] $}
		\FOR{$i_x = 1, \cdots, d_x$}
		\IF{$i_x \notin \Gamma_x$}
		\STATE{$S.\text{enqueue}(i_x)$}
		\STATE{Set $Pr.\text{enqueue}(\sum_{i_y,i_z}r(i_x, i_y, i_z| \Upsilon, F, \Gamma ))$}
		\STATE{Set $Pr_{\text{sum}} = Pr_{\text{sum}} + \sum_{i_y,i_z}r(i_x, i_y, i_z| \Upsilon, F, \Gamma )$}				
		\ENDIF
		\ENDFOR
		\IF{$Pr.\text{isempty}()$ or $Pr_{\text{sum}} = 0$}
		\STATE{return $\{ \}$}
		\ENDIF		
		\STATE{$\Gamma_x^{\prime} = \text{random.sample}(S, n=\tau, p=Pr / Pr_{\text{sum}})$}
	\end{algorithmic}
	\label{algo:propose_slice_to_label}
\end{algorithm}
We first compute a residual map $r$ indicating the likelihood of which slices to label in the next round. 
\begin{align}
	r(\mathbf{x}) = \omega_1 d_1(\Upsilon(\mathbf{x}), F(\mathbf{x})) + (1 - \omega_1) d_2(\mathbf{x}, \Gamma)
	\label{eq:residule_map}
\end{align}
We consider two factors in equation~\ref{eq:residule_map}: the difference between $\Upsilon, F$ and the closeness to already labeled slices $\Gamma$; $d_1, d_2$ are distance/dis-similarity metrics, and $\omega_1$ is the weight to balance these two factors. Conceptually $r$ works the opposite way to $\mathbf{W}$ in equation~\ref{eq:spatial_weight_forms_1}: regions with lower confidence and higher (approximate) segmentation errors are more likely to be sampled. Instead of selecting slices according to ordered magnitudes of $r$, we introduce randomness in step 15 to balance optimization objective and spatial uniformity. This helps for bias reduction in equation~\ref{eq:spatial_weighted_loss} when the weights. Eventually, we could substitute the user decision process (step 5 in algorithm~\ref{algo:interactive_learning}) with the model guided proposals. Figure~\ref{fig:flowchart_inter} gives an illustration of whole process of interactive segmentation with model guidance. 

The input $F$ in algorithm~\ref{algo:propose_slice_to_label} could be either pre-trained on another set of samples (warm-up case) or totally untrained (cold-start case). Though generally a pre-trained $F$ could perform better in providing proposals, we will show that quite surprisingly, an untrained $F$ could quickly learn to provide effective guidance after a few rounds of interactions. 

% In reality, we have neither of the above two assumtions, however, an increamental model guided labeling scheme could be inspired and merged into Algorithm [], to solve the problems in above settings in an alternating way, see Algorithm [] below. In particular, rather than directly solving equation~\ref{eq:opt_search_slice_incremental}, we propose an randomized greedy method in Algorithm [], without needing to actually propagaging candidate labeled slices to 3D, which balances the optimization target and spatial uniformity for bias reduction.

\subsection{Dynamic Interactive Learning}
\label{sec:dynam_inter_learn}

Consider the practical cases where we continuously have new images arriving to be segmented. In addition to producing accurate masks for arriving images, we also need a continuously improving automated segmenter that yields robust predictions on unseen images. Equipped with modules discussed previously, we now have an integrated solution with reduced user labor. 

Inherently, the images and proxy masks constitute a sequence of training samples with weak supervision: $T=\{(X_i, \Upsilon_i^j)\}$ where $i, j$ represent image indices and steps of interactive segmentation. The segmenter could be trained dynamically following section~\ref{sec:dynam_learn}. Comparing to strong supervision, some subtle adjustments should be applied to retain comparable performances, considering that the proxy mask at earlier interaction steps could carry more noises that disrupt training. The integrated learning process is detailed in algorithm~\ref{algo:dynam_inter_learn}. 

\begin{algorithm}
	\caption{Dynamic Interactive Training}
	\begin{algorithmic}[1]
		\STATE{\textbf{Input}: Image stream $T=\{X_k\}$, $\gamma$, $n_{\text{inter}}$, $b$, $s$}
		\STATE{\textbf{Output}: Proxy masks $\{\Upsilon_k\}$, $F$}
		\STATE{Set $t=0, B=\{ \}, S=\{ \}, \text{isFirst = 1}, F_{\text{prev}}={\mathbf{x} \mapsto 0}$}
		\WHILE{not $T.\text{isempty}()$} 
		\STATE{Set $X=T.\text{dequeue}()$}				
		\STATE{Set $\Gamma = \{ \}$, $i = 0$, $\Upsilon=\mathbf{0}$}
		\WHILE{$|\Gamma| < \gamma$ and $i < n_{\text{inter}}$}
		\IF{$\text{isFirst}$} 
		\STATE{Set $\Gamma^{\prime} = \text{random.sample}()$}					
		\ELSE
		\STATE{Set $\Gamma^{\prime}$ (algorithm~\ref{algo:propose_slice_to_label})}					
		\ENDIF
		\STATE{\textbf{User}: Draw labels for slices in $\Gamma^{\prime}$}
		\STATE{Set $\Gamma = \Gamma \cup \Gamma^{\prime}$}								
		\STATE{Set $\Upsilon$ with $\Gamma^{\prime}$ (algorithm~\ref{algo:slice_prop_register})}
		\STATE{Set $\mathbf{W}, \lambda_L(t), \lambda_Y(t), n_F(i)$}
		\STATE{Set $\tilde{\Upsilon} =  (1 - \lambda_Y(t))F_{\text{prev}}(X) + \lambda_Y(t)\Upsilon$}
		\FOR{$j = 1, \cdots, n_F(i)$}
		\STATE{Set $\theta = \theta - \eta \bigtriangledown_{\theta} L(\theta | \{(X, \tilde{\Upsilon})\} \cup S)$}
		\ENDFOR								
		\STATE{Set $B, S$ with $(X, \tilde{\Upsilon})$ (algorithm~\ref{algo:buffer_update})}								
		\IF{$\text{isFirst}$} 
		\STATE{Set $\text{isFirst = 0}$}
		\ENDIF								
		\STATE{Set $i = i + 1$}
		\ENDWHILE
		\STATE{Set $F_{\text{prev}} = F$}
		\STATE{Set $t = t + 1$}
		\ENDWHILE
	\end{algorithmic}
	\label{algo:dynam_inter_learn}
\end{algorithm}

Note that we activate model guided annotation after the segmenter finished at least one round of training (step 11). Also, we fix the previous segmenter when computing smoothed label inside of the interactive segmentation loop (step 17). As the system sees more samples, the trained segmenter is more capable of generating robust predictions on unseen cases as well as more precise guidance for user interventions.

\section{Experiments and Results}\label{sec:exp}

%The most popular metric for segmentation performance is the Dice coefficient: $\text{Dice}(\mathcal{R}_{\text{true}}, \mathcal{R}_{F}) = 2|\mathcal{R}_{\text{true}} \cap \mathcal{R}_{F}| / (|\mathcal{R}_{\text{true}}| + |\mathcal{R}_{F}|)$ where $\mathcal{R}_{\text{true}}$ and $\mathcal{R}_{F}$ represents the foreground regions segmented by ground truth and segmenter $F$ respectively. The original output of $F$ is a probability map and a cut-off threshold (commonly $0.5$) could be used to produce $\mathcal{R}_{F}$. A soft version of Dice score [35] is therefore also used as both evaluation metric and loss function: $\text{Dice}_{\text{soft}} = 2\| Y \odot p \|_1 / (\|Y\|_1 + \|p\|_1)$ where $Y$ is the true label, $p=F(X)$ is the predicted probability map, and $\odot$ represents element-wise multiplication. 

The most popular evaluation metric for segmentation is the Dice coefficient. In practical clinical applications, it is critical to evaluate the resources (user time, in particular) needed to achieve an acceptable accuracy, rather than simply competing on dice scores assuming unlimited resources. With physicians in the loop (algorithm~\ref{algo:interactive_learning}), we could measure the real time spent on labeling/decision. However, this leads to significant human resources and difficulty to control variations among different physicians. We therefore simulate user behaviors in a controlled way for consistent comparisons across different settings and with other works. Authors in~\cite{ref_23} proposed two types of simulated user inputs: good and bad interactions. They assumed the simulated ``users'' could instantly pick the optimal locations for correction clicks based on the gap between ground truth and model predictions to form ``good'' interactions; on the other hand, ``bad'' interaction means randomly clicks inside the image domain. Obviously, the availability of ground truth is a strong assumption and zero time overheads for decision is also unrealistic. In contrast, we are more interested in offering complete guidance for user inputs without knowing the ground truth. Also, we want to guarantee that even "bad" interactions won't hurt the overall performance. Our experiments focus on the following three kinds of evaluations:

\textbf{(A)} Dice scores on a held-out set. It is obvious the upper bound of dice score will be achieved when training the model in an offline way with ground truth available; therefore we won't expect our framework to outperform this benchmark. Rather, we show that our framework could approach this benchmark and discuss optimal strategies.

\textbf{(B)} Dice scores vs. the amount of user labors on a held-out set in three different cases.

\textbf{(B-1)} Cold-start case. For each image to be segmented, the underlying segmenter is initiated without any pre-training.

\textbf{(B-2)} Warm-up case. The segmenter was pre-trained on an offline dataset before used for guided interactions. 

\textbf{(B-3)} Dynamic evaluation. Instead of treating each image as isolated ones, we have them arriving sequentially to our system. This is a more practical case where we want to investigate how our system could learn from experiences of segmenting past samples.

\textbf{(C)} Minimum user labor needed to reach a desired dice score. This is the major concern in deploying our system in clinical applications. We set a threshold for dice score, with samples entering our framework, we keep monitoring the minimum user labor needed in interactive learning algorithm to reach this threshold.

Note that with same resources, there could be two different types of segmentation predictions generated with our framework: the proxy mask $\Upsilon$ and model prediction $F(X)$. It is reasonable to use $\Upsilon$ when interactive process is involved for the test image and $F(X)$ otherwise. In subsequent sections, we will thoroughly evaluate our system according to the above perspectives. 

\subsection{Datasets}\label{sec:exp:datasets}

We use two data sets in the experiments, BraTS2015~\cite{ref_57} and NCI-ISBI2013~\cite{ref_59} with identical data splitting settings as in~\cite{ref_23}:

(1) Brain Tumor Segmentation Challenge 2015 (BraTS2015)~\cite{ref_23, ref_57}. The target is to segment whole brain tumor in 274 FLAIR images and we set $N_{\text{train1}} = N_{\text{train2}} = 117, N_{\text{test}} = 40$.

(2) NCI-ISBI2013~\cite{ref_23, ref_59} include 80 samples for automated segmentation of prostate structures. We set $N_{\text{train1}}=N_{\text{train2}}=32, N_{\text{test}}=16$.

Different from~\cite{ref_23}, we combine $N_{\text{train1}}$ and $N_{\text{train2}}$ for offline/dynamic training and use $N_{\text{test}}$ for held-out evaluations. Similarly, we first normalize all images by mean/std of the whole data set, crop each image by its non-zero boundaries with a margin of $[0, 10]$ pixels, and resize to $55\times 55\times 30$. During training, we apply random flipping and rotation on the fly as data augmentation. 

\subsection{Dynamic learning}\label{sec:exp:dynamic_learning}

\begin{table*}[htb]
	\caption{DICE scores on $D_{eval}$. RP and LS represents applying replay and label smoothing techniques respectively}
	\label{tab:dice_dynamic_brats_nci}
	\centering
	\begin{tabular}{lclclclcl}
		\hline
		BraTS2015 & mean  & std & min & max \\
		\hline
		Offline    & 0.79$\pm$0.0035 & 0.12$\pm$0.0051 & 0.46$\pm$0.06 & 0.92$\pm$0.0025 \\
		Dynamic   & 0.72$\pm$0.0071  & 0.16$\pm$0.0067  & 0.35$\pm$0.052   & 0.90$\pm$0.0042 \\
		Dynamic+RP   & 0.79$\pm$0.0039 & 0.12$\pm$0.0052  & 0.46$\pm$0.045 & 0.93$\pm$0.0045  \\
		Dynamic+LS    & 0.72$\pm$0.0028  & 0.16$\pm$0.0046    &  0.34$\pm$0.030   & 0.89$\pm$0.0095   \\
		Dynamic+RP+LS   & 0.79$\pm$0.0028   & 0.13$\pm$0.0033   & 0.46$\pm$0.038   & 0.93$\pm$0.005   \\
		\hline
		NCI-ISBI2013 & mean  & std & min & max \\
		\hline
		Offline       & 0.80$\pm$0.0052  & 0.046$\pm$0.0019  & 0.70$\pm$0.02 & 0.87$\pm$0.0083 \\
		Dynamic         & 0.71$\pm$0.0051  & 0.066$\pm$0.0041  & 0.58$\pm$ 0.016 & 0.80$\pm$0.011 \\
		Dynamic+RP       & 0.82$\pm$0.0041  & 0.039$\pm$0.0024 & 0.72$\pm$0.0151  & 0.88$\pm$0.0058 \\
		Dynamic+LS    & 0.70$\pm$0.0083 & 0.078$\pm$0.025  & 0.53$\pm$0.099  & 0.80$\pm$0.021 \\
		Dynamic+RP+ LS   & 0.82$\pm$0.0036   & 0.039$\pm$0.0037  & 0.73$\pm$0.017  & 0.88$\pm$0.0079 \\
		\hline
	\end{tabular}	
\end{table*}

\begin{table*}[htb]
	\caption{Dice scores evaluated on $D_{eval}$/$\tilde{T}^{t}_{train}$ vs. number of first $t$ samples in $T_{train}$ participated in training}
	\label{tab:dice_dynamic_seqeval_brats_nci}
	\centering
	\begin{tabular}{lclclclclclc|}
		\hline
		& \multicolumn{6}{c}{BraTS2015, avg. dice on $D_{eval}$/$\tilde{T}^{t}_{train}$} \\
		\cmidrule{2-7}
		$\#$ samples in training   & &40  & 80  & 120  & 160 & 200  \\
		\hline
		Offline     & & 0.69/0.72    & 0.76/0.77     & 0.78/0.79     & 0.78/0.79      & 0.78/0.80   \\
		Dynamic     & & 0.69/0.63    & 0.70/0.69    & 0.71/0.71   & 0.71/0.71     & 0.72/0.72    \\  
		Dynamic+RP  & & 0.77/0.78    & 0.79/0.79    & 0.79/0.80    &  0.79/0.80   & 0.79/0.80    \\
		Dynamic+LS   &  & 0.69/0.63    & 0.70/0.69     & 0.71/0.71     & 0.72/0.71     & 0.72/0.71       \\
		Dynamic+RP+LS &   & 0.77/0.78   & 0.79/0.79    & 0.79/0.80     & 0.79/0.80     & 0.79/0.80       \\
		\hline
		& \multicolumn{6}{c}{NCI-ISBI2013, avg. dice on $D_{eval}$/$\tilde{T}^{t}_{train}$} \\
		\cmidrule{2-7}
		$\#$ samples in training   &10  & 20  & 30  & 40 & 50 & 60 \\
		\hline
		Offline     &  0.73/0.72   &   0.76/0.74  & 0.78/0.77  & 0.78/0.78   &  0.79/0.79  & 0.80/0.81   \\
		Dynamic     & 0.59/0.41     &  0.65/0.58  & 0.68/0.59  & 0.68/0.64   &  0.70/0.56 & 0.70/0.62   \\
		Dynamic+RP   & 0.74/0.70   & 0.78/0.75 &  0.79/0.76  & 0.81/0.79 & 0.81/0.80 & 0.82/0.82   \\
		Dynamic+LS     & 0.60/0.40   & 0.65/0.60 & 0.67/0.60 & 0.69/0.66  & 0.69/0.55  & 0.70/0.64 \\
		Dynamic+RP+LS    & 0.74/0.69    & 0.78/0.74  & 0.79/0.76  & 0.81/0.79   & 0.81/0.80  & 0.81/0.82 \\
		\hline
	\end{tabular}	
\end{table*}

We evaluate the dynamic learning module in section~\ref{sec:dynam_learn}. Instead of a complete discussion of all OCL tricks~\cite{ref_25}, we show that our dynamic learning performances could match the offline benchmark. Let $T_{train}=\{(X_1, Y_1), \dots, (X_{N_{train}}, Y_{N_{train}})\}$ where $N_{train} = N_{train1} + N_{train2}$ and $D_{eval}$ be $N_{test}$ evaluation samples. As we take more training samples from $T_{train}$, we compare (A) dynamic learning and (B) offline benchmark on both the unseen samples and the held-out set. We repeat this process with randomly reordered $T_{train}$, see algorithm~\ref{algo:eval_dynamic_learning}.

\begin{algorithm}
	\caption{Evaluate dynamic learning}
	\begin{algorithmic}[1]
		\STATE{\text{Input}: $T_{train}$, $D_{eval}$, $N_{perm}$, $F$}
		\FOR{$i = 1, \dots, N_{perm}$}
		\STATE{Set $T_{train} \leftarrow \text{random.permutation}(T_{train})$}
		\FOR{$t = 1, \dots, N_{train}$}				
		\STATE{Set $T_{train}^t = \{(X_1, Y_1), \dots, (X_t, Y_t)\}$}
		\STATE{Set $\tilde{T}_{train}^t=T_{train} \setminus T_{train}^t$}
		\STATE{\textbf{A: Offline benchmark}:  train $F$ offline on $T_{train}^t$}
		\STATE{\textbf{B: Dynamic training}:  train $F$ dynamically on $T_{train}^t$}
		\STATE{Evaluate \textbf{A}, \textbf{B} on $D_{eval}$ and $\tilde{T}_{train}^t$}
		\ENDFOR
		\ENDFOR
		\STATE{Aggregate statistics}
	\end{algorithmic}
	\label{algo:eval_dynamic_learning}
\end{algorithm}

\begin{figure}
	\centering
	\begin{subfigure}[b]{\textwidth}
		\includegraphics[width=\textwidth]{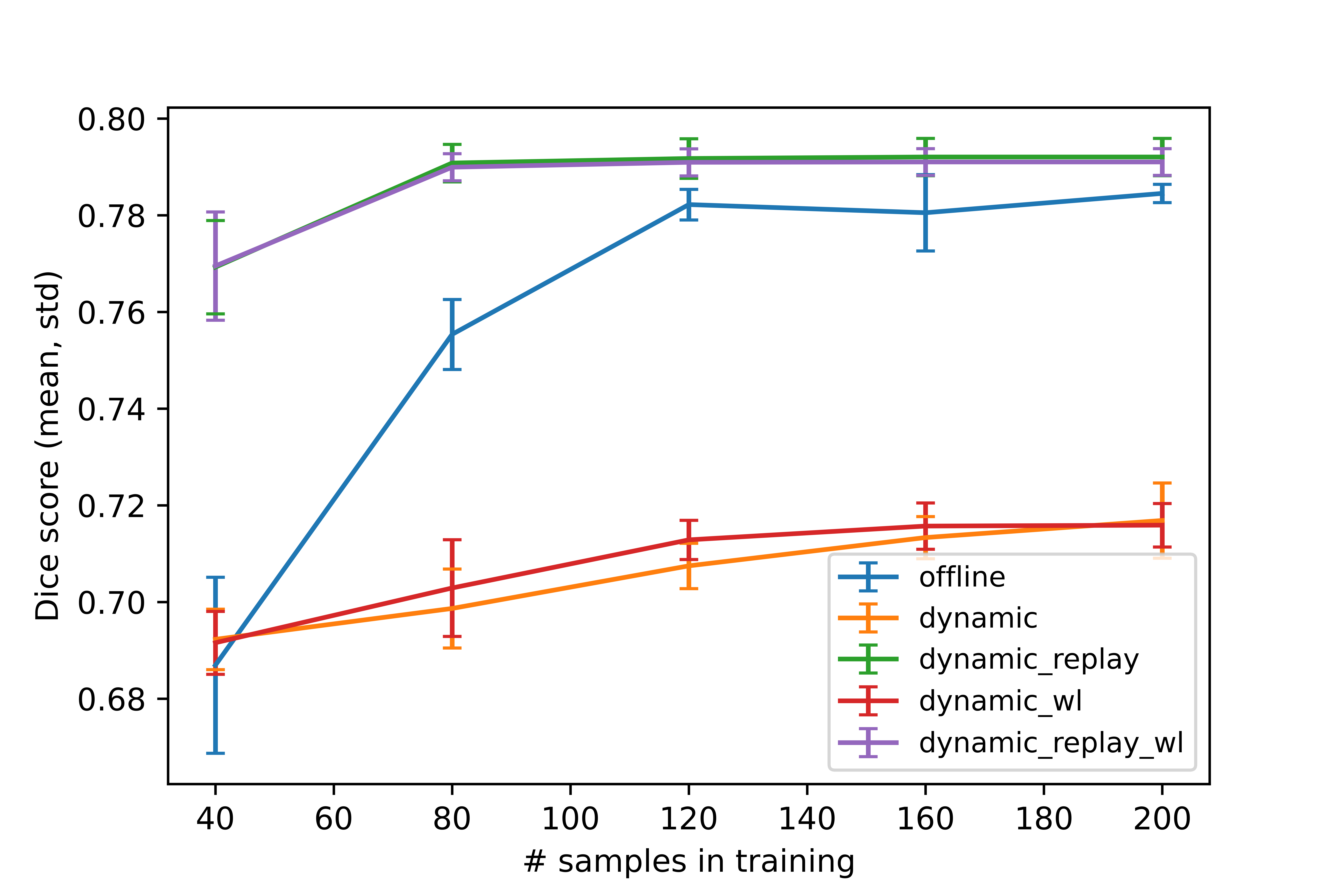}
	\end{subfigure}
	\begin{subfigure}[b]{\textwidth}
		\includegraphics[width=\textwidth]{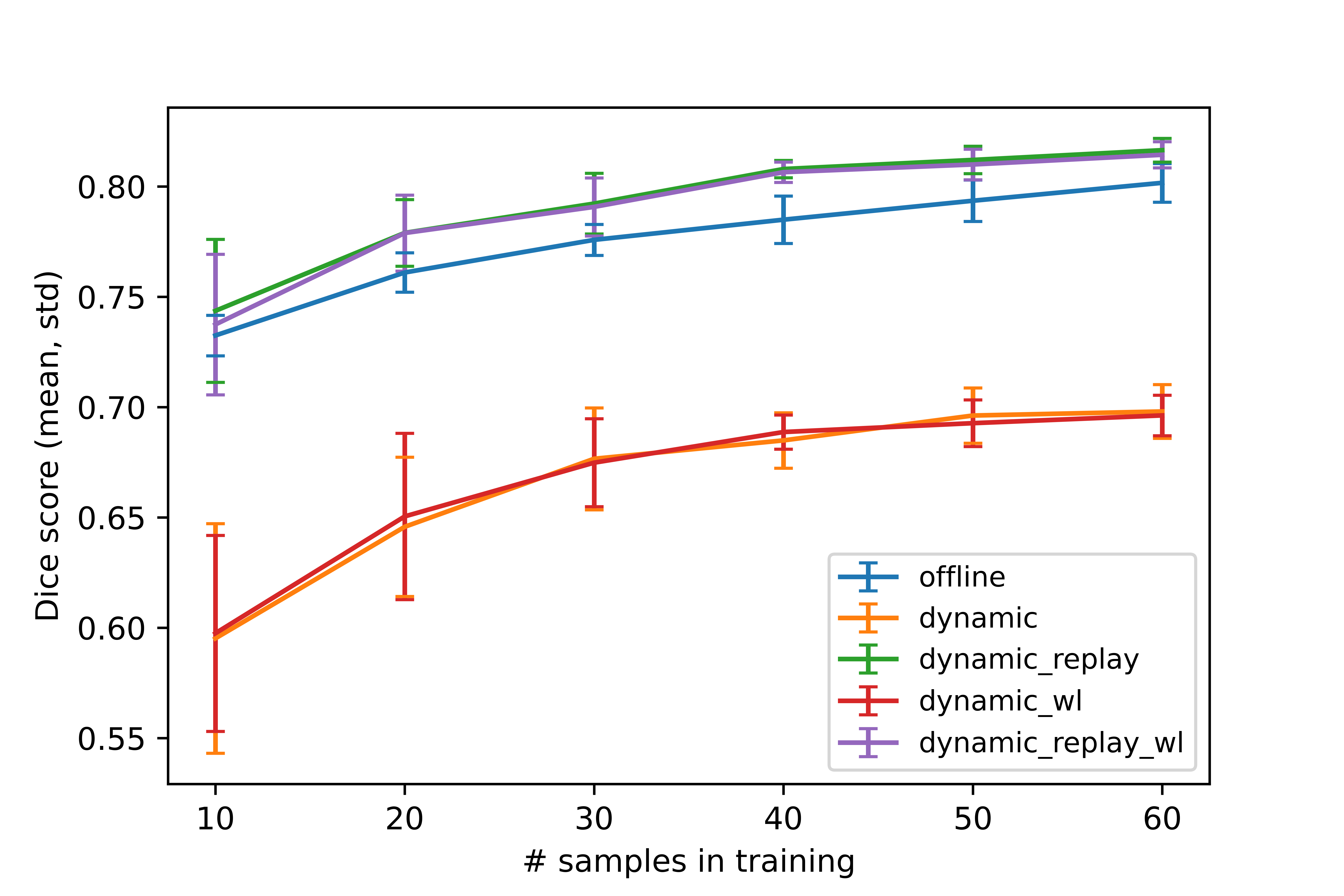}
	\end{subfigure}
	\caption{BraTS2015 (top), NCI-ISBI2013 (bottom). Average Dice scores on $D_{eval}$ vs. number of samples participated in offline/dynamic training. The error bar (std) came from repeated runs of random permutations of $T_{train}$, shown by curves}
	\label{fig:dice_dynamic_seqeval_brats_nci_eval}
\end{figure}

Table~\ref{tab:dice_dynamic_brats_nci} provided the dice scores $D_{eval}$ across different settings. The numbers appending $\pm$ is the std coming from multiple permutations of $T_{train}$. We see that feeding samples sequentially to model disturbs the training process, but our replay scheme (algorithm~\ref{algo:buffer_update}) helped to mitigate catastrophic forgetting and matched offline benchmark. Label smoothing does not lead to further improvement, since past experiences from the segmenter won't bring additional knowledge with the existence of ground truth. Table~\ref{tab:dice_dynamic_seqeval_brats_nci} shows the trending patterns as we gradually take more samples for training. The dice scores were collected from both $D_{eval}$ and $\tilde{T}^{t}_{train}$. Similarly, our replay technique significantly helped dynamic learning. In addition, figure~\ref{fig:dice_dynamic_seqeval_brats_nci_eval} straightforwardly shows that dynamic learning could even produce more robust results than offline benchmark; part of the reasons may be that the data evolution mechanism prevents frequent revisits to the same sample and thus mitigates over fitting, for the case of small sample size. As more samples are involved, the performances of offline benchmark and dynamic learning become closer.

\subsection{User Interactions and Model Guidance}\label{sec:exp:user_interaction}

We investigate user interaction and model guided annotation modules (Algorithms~\ref{algo:interactive_learning} and~\ref{algo:propose_slice_to_label}) under both cold-start (untrained $F$) and warm-up (pre-trained $F$ on $T_{train}$) conditions. Also, both stand-alone and dynamic evaluations will be performed on $D_{eval}$. We compare the results with/without model guided annotations (algorithm~\ref{algo:propose_slice_to_label}) and weighted loss strategy (equation~\ref{eq:spatial_weighted_loss}).

It is difficult to have a fully fair comparison with existing works on deep learning based interactive segmentation. The closest settings we found is~\cite{ref_23} that showed the changes of performances vs. the amount of simulated user labors (number of clicks). However, as noted above, they used user inputs to generate hint maps as auxiliary input; and the identical process should be executed during training with ground truth labels provided. On one hand, they showed outstanding dice scores with just 25 user clicks; on the other hand, their performance heavily relies on the alignment between training and testing procedures. In reality, we cannot expect to reproduce their learning process due to the immediate lack of ground truth labels. However, our system is flexible enough to produce reasonable accuracy following similar testing-stage interaction process. In each step, we allow the simulated user to delineate one slice of the target on coronal and sagittal planes each, and then the proxy mask (algorithm~\ref{algo:slice_prop_register}) will be used as predicted segmentation. One ``click'' seems cheaper than drawing a slice boundary, but it is hard to quantify and compare total user labors. With efficient tools~\cite{ref_61}, the drawing boundaries could be much faster. Therefore, we will list our results side by side with the numbers from~\cite{ref_23} to form a straightforward comparison of the trend, rather than a strict competition on user time.

In tables~\ref{tab:brats_interactive_compare_rl} and~\ref{tab:nci_interactive}, we list dice scores on $D_{eval}$ for certain rounds of interactions. In each round,``user'' selects one slice from each along coronal and sagittal planes to label. By default we assum ``bad interaction'' in~\cite{ref_23} where slices are selected randomly. With model guidance (MG) enabled, our system directly determines which slices to label. From table~\ref{tab:brats_interactive_compare_rl}, we see there is really no ``bad interactions'' with our framework comparing to~\cite{ref_23}. With randomly selected annotation locations, \cite{ref_23} presents decaying accuracies as more user inputs are collected. In contrast, our system showed increasing accuracy which eventually matches the state-of-the-arts number in~\cite{ref_23}. This observation demonstrates the robustness of our system to user actions.

In addition, model guidance (MG) greatly improved the performances during first few steps of interactions. Moreover, a pre-training segmenter (warm-up) does help during early stages but the improvement is not as large as that brought by MG. Even without pre-training (cold-start), the system could produce better performance than Min-Cut~\cite{ref_59, ref_23, ref_17} at step 3 with only $\sim$27\% of the total human efforts required for full annotation (``full efforts''). In table~\ref{tab:nci_interactive}, we have consistent observations that both MG and warm-up started to help when user inputs started to reach $\sim$24\% of the ``full efforts'' at step 3; also, weighted loss strategy (equation~\ref{eq:spatial_weighted_loss}) further improved the min dice scores. Unlike~\cite{ref_23}, the full ground truth labels are not needed throughout interactive learning.

\begin{table*}[htb]
	\caption{\textbf{BraTS2015}. Interactive segmentation vs. user labors, comparing with tables 2 and 6 in~\cite{ref_23}. MG means to apply model guided annotation (algorithm~\ref{algo:propose_slice_to_label}) and WL means to apply weighted loss (equation~\ref{eq:spatial_weighted_loss}); C+S means number of (target) slices from coronal and sagittal planes.}
	\label{tab:brats_interactive_compare_rl}
	\centering
	\resizebox{\textwidth}{!}{%
	\begin{tabular}{lclclclclcl}
		\hline
		& \multicolumn{5}{c}{Dice score: mean[min (if provided), max (if provided)]} \\
		\cmidrule{2-6}
		$\#$ Steps   &1  & 2  & 3  & 4 & 5 \\
		$\#$ Annotated Points [23]  &5  & 10  & 15  & 20 & 25  \\		
		\hline
		Min-Cut~\cite{ref_23}      &  79.52   & 79.97  & 80.22   & 80.46  &  80.69   \\
		IteR-MRL goodInter~\cite{ref_23}  & 84.35 & 86.78 & 87.61 & 88.18 & \textbf{88.53}  \\
		IteR-MRL noInter~\cite{ref_23}  & 78.60 & 79.53 & 80.15 & 80.56 & 80.78  \\
		IteR-MRL badInter~\cite{ref_23} & 76.86 & 75.47 & 74.84 & 74.29 & 72.76  \\
		\hline
		$\#$ Slices delineated (C+S) &2+2 &3+3 &4+4 &5+5 &6+6  \\
		$\#$ Mean full target slices (C/S) &23/43 &23/43 &23/43 &23/43 &23/43  \\
		\hline
		cold-start &0.60[0.11, 0.89]  &0.76[0.17, 0.92] &0.79[0.20, 0.94] &0.83[0.22, 0.95] &0.87[0.31, 0.95]  \\
		cold-start MG &0.76[0.17, 0.92]  &0.79[0.25, 0.92]  &0.82[0.41, 0.93]  &0.86[0.65, 0.94]  &0.87[0.69, 0.94]  \\
		cold-start MG + WL &0.77[0.16, 0.93]  &0.79[0.24, 0.91]  &0.82[0.39, 0.93]  &0.86[0.64, 0.94]  &0.87[0.69, 0.94]  \\
		warmup MG &0.79[0.40, 0.92]  &0.82[0.53, 0.93]  &0.84[0.62, 0.92]  &0.86[0.66, 0.94]  &0.88[0.71, 0.94]  \\
		warmup MG + WL &0.79[0.36, 0.92]  &0.83[0.56, 0.92]  &0.85[0.62, 0.93]  &0.86[0.64, 0.94]  &0.88[0.71, 0.95]  \\
		\hline
	\end{tabular}}	
\end{table*}

\begin{table*}[htb]
	\caption{\textbf{NCI-ISBI2013}. Interactive segmentation vs. user labors. MG means to apply model guided annotation (algorithm~\ref{algo:propose_slice_to_label}) and WL means to apply weighted loss (equation~\ref{eq:spatial_weighted_loss}); C+S means number of (target) slices from coronal and sagittal planes.}
	\label{tab:nci_interactive}
	\centering
	\resizebox{\textwidth}{!}{%
	\begin{tabular}{lclclclclclc|}
		\hline
		& \multicolumn{6}{c}{Dice score: mean[min, max]} \\
		\cmidrule{2-7}
		$\#$ Steps   &1  & 2  & 3  & 4 & 5 & 6 \\
		$\#$ Annotated Slices (C+S)   &1+1  & 2+2 & 3+3  & 4+4 & 5+5 & 6+6 \\
		$\#$ Mean full target slices (C/S) &17/45 &17/45 &17/45 &17/45 &17/45 &17/45 \\
		\hline
		cold-start      &  0.54[0.073, 0.78]   & 0.65[0.46, 0.79]   &   0.71[0.57, 0.84]   & 0.78[0.60, 0.88]   &  0.82[0.75, 0.91] & 0.84[0.76, 0.92]   \\
		cold-start MG     & 0.53[0.076, 0.77]     &  0.73[0.60, 0.87]  & 0.75[0.64, 0.85] & 0.76[0.69, 0.87]   &  0.81[0.73, 0.88]  & 0.83[0.70, 0.90]   \\
		cold-start MG + WL    & 0.53[0.07, 0.76]   & 0.72[0.57, 0.86] &  0.75[0.65, 0.86]  & 0.77[0.69, 0.88] & 0.81[0.74, 0.88]  & 0.84[0.76, 0.92]   \\
		warmup MG     & 0.54[0.071, 0.76]    & 0.71[0.49, 0.84] & 0.75[0.50, 0.87] & 0.79[0.68, 0.88]  & 0.82[0.76, 0.90]  & 0.85[0.78, 0.91]  \\
		warmup MG + WL    & 0.54[0.073, 0.77]    & 0.71[0.51, 0.85]   & 0.76[0.52, 0.87]  & 0.79[0.70, 0.86]   & 0.82[0.75, 0.91]  & 0.86[0.77, 0.93] \\
		\hline
	\end{tabular}}	
\end{table*}

\subsection{Dynamic learning with Interactive Segmentation}\label{sec:exp:dynam_learn_with_inter}

We discussed dynamic learning and interactive segmentation modules separately. We assumed the existence of ground truth labels in dynamic learning and warm-up stage of interactive segmentation, to explicitly analyze the properties of proposed strategies including replay, label smoothing, model guided interaction and weighted loss. Now we investigate our integrated dynamic interactive learning framework.

First, instead of treating $D_{eval}$ as individual query images in section~\ref{sec:exp:user_interaction}, we align them in a queue and allow segmenter to learn from past annotations (dynamic evaluation). Table~\ref{tab:brats_nci_interactive_dynameval} listed the best results from tables~\ref{tab:nci_interactive} and~\ref{tab:brats_interactive_compare_rl} along with dynamic evaluation. For NCI-ISBI2013, evaluating $D_{eval}$ dynamically does not help to boost up the results further, but for BraTS2015, it shows improvements under cold-start setting. The evaluation set for NCI-ISBI2013 has 16 images while BraTS2015 has 40 images. Comparing to model guided interaction that instantly provides improvements on a single image, dynamic learning requires longer sequence to stabilize and reveal its advantage. Also, in warm-up settings there is already a mature segmenter so dynamic evaluation does not bring additional gain.

\begin{table*}[htb]
	\caption{Interactive segmentation vs. user labors. Compare static vs. dynamic evaluation. dynam(+RP) means to allow segmenter to learn from past annotations by treating $D_{eval}$ as a stream.}
	\label{tab:brats_nci_interactive_dynameval}
	\centering
	\resizebox{\textwidth}{!}{%
	\begin{tabular}{lclclclclclc|}
		\hline
		& \multicolumn{6}{c}{BraTS2015, Dice score: mean[min, max]} \\
		\cmidrule{2-7}
		$\#$ Steps   &1  & 2  & 3  & 4 & 5 & 6 \\
		$\#$ Annotated Slices (C+S)   &1+1  & 2+2 & 3+3  & 4+4 & 5+5 & 6+6 \\
		$\#$ Mean full target slices (C/S) &23/43 &23/43 &23/43 &23/43 &23/43 &23/43 \\
		\hline
		cold-start(MG+WL)    &0.48[0.03, 0.87]   & 0.77[0.16, 0.93]  & 0.79[0.24, 0.91] & 0.82[0.39, 0.93]  & 0.86[0.64, 0.94] & 0.87[0.69, 0.94]  \\
		cold-start(MG+WL) dynam(+RP) & 0.48[0.03, 0.90]  & 0.77[0.18, 0.92]  & 0.82[0.57, 0.92]   & 0.84[0.62, 0.93]  & 0.86[0.66, 0.94]  & 0.88[0.69, 0.95]  \\
		warmup(MG+WL)    & 0.48[0.02, 0.88]  & 0.79[0.36, 0.92]  & 0.83[0.56, 0.92]  & 0.85[0.62, 0.93]  & 0.86[0.64, 0.94]   & 0.88[0.71, 0.95]  \\
		warmup(MG+WL) dynam(+RP)    & 0.48[0.03, 0.88]  & 0.79[0.39, 0.92]  & 0.83[0.54, 0.93]  & 0.85[0.62, 0.95]  & 0.86[0.65, 0.95]  &  0.88[0.68, 0.95]    \\		
		\hline
		& \multicolumn{6}{c}{NCI-ISBI2013, Dice score: mean[min, max]} \\
		\cmidrule{2-7}
		$\#$ Steps   &1  & 2  & 3  & 4 & 5 & 6 \\
		$\#$ Annotated Slices (C+S)   &1+1  & 2+2 & 3+3  & 4+4 & 5+5 & 6+6 \\
		$\#$ Mean full target slices (C/S) &17/45 &17/45 &17/45 &17/45 &17/45 &17/45 \\
		\hline
		cold-start(MG+WL)    & 0.53[0.07, 0.76]   & 0.72[0.57, 0.86] &  0.75[0.65, 0.86]  & 0.77[0.69, 0.88] & 0.81[0.74, 0.88]  & 0.84[0.76, 0.92]   \\
		cold-start(MG+WL) dynam(+RP) & 0.53[0.07, 0.79]  & 0.71[0.52, 0.86]  & 0.74[0.55, 0.87]  & 0.79[0.68, 0.90]  & 0.81[0.75, 0.91]  & 0.84[0.77, 0.91]  \\
		warmup(MG+WL)    & 0.54[0.073, 0.77]    & 0.71[0.51, 0.85]   & 0.76[0.52, 0.87]  & 0.79[0.70, 0.86]   & 0.82[0.75, 0.91]  & 0.86[0.77, 0.93] \\
		warmup(MG+WL) dynam(+RP)    & 0.54[0.07, 0.79]     & 0.71[0.51, 0.86]   & 0.76[0.61, 0.89]  & 0.79[0.68, 0.88]  & 0.82[0.75, 0.90]   & 0.85[0.78, 0.91]    \\
		\hline
	\end{tabular}}	
\end{table*}

\begin{figure*}[htb]
	\centering
	\includegraphics[width=\textwidth]{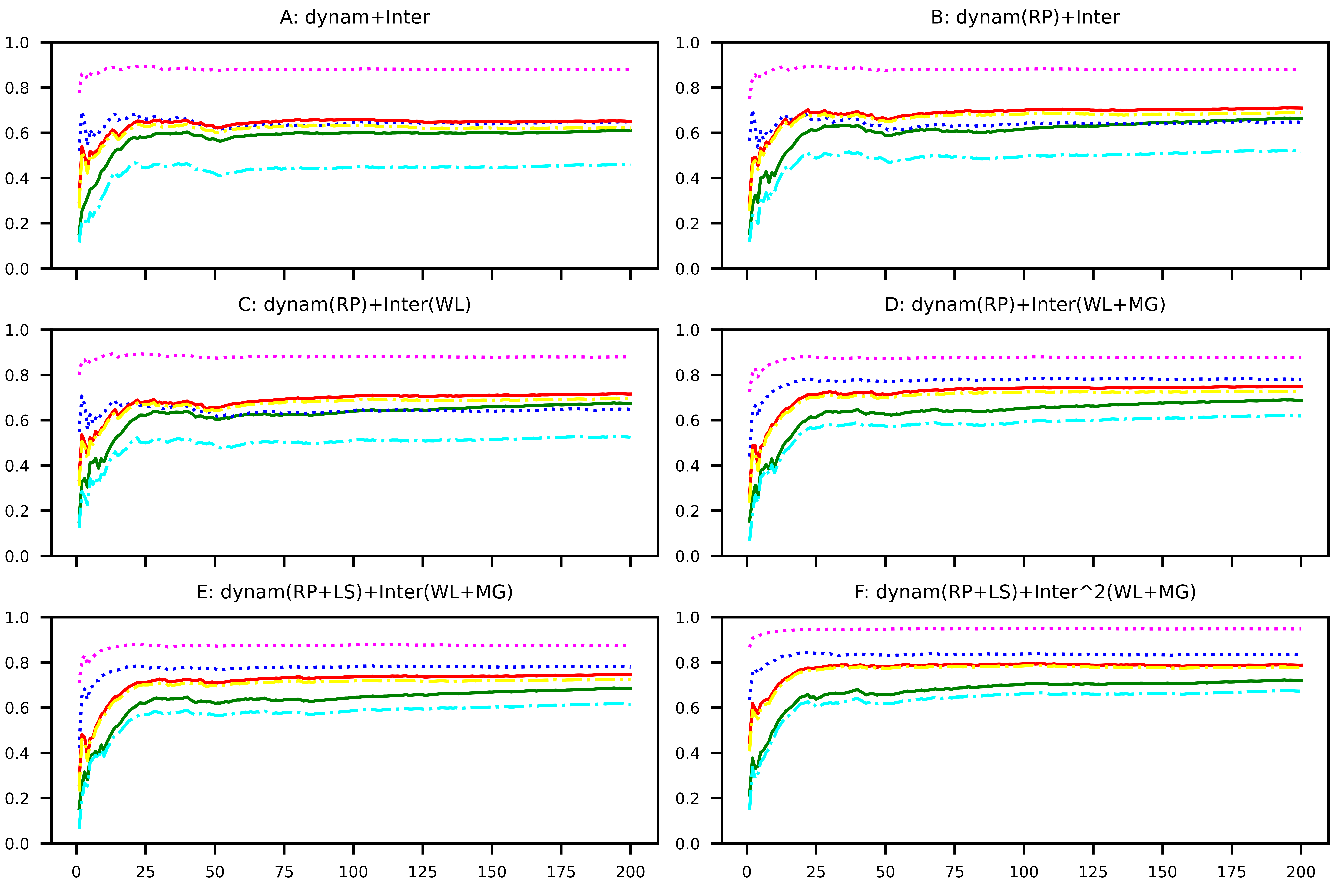}
	\caption{BraTS2015, Cum. avg. dice scores vs. number of samples in dynamic interactive training. ``Dice\_Y\_vs\_F'': $Dice(Y, F(X))$,  ``Dice\_Y\_vs\_Yp'': $Dice(Y, \Upsilon)$, ``Dice\_Yp\_vs\_F'': $Dice(\Upsilon, F(X))$; round $t$ means the $t^{\text{th}}$ round of interactive learning.}
	\label{fig:dice_triangle_brats_nci_brats}
\end{figure*}

\begin{figure*}[htb]
	\centering
	\includegraphics[width=\textwidth]{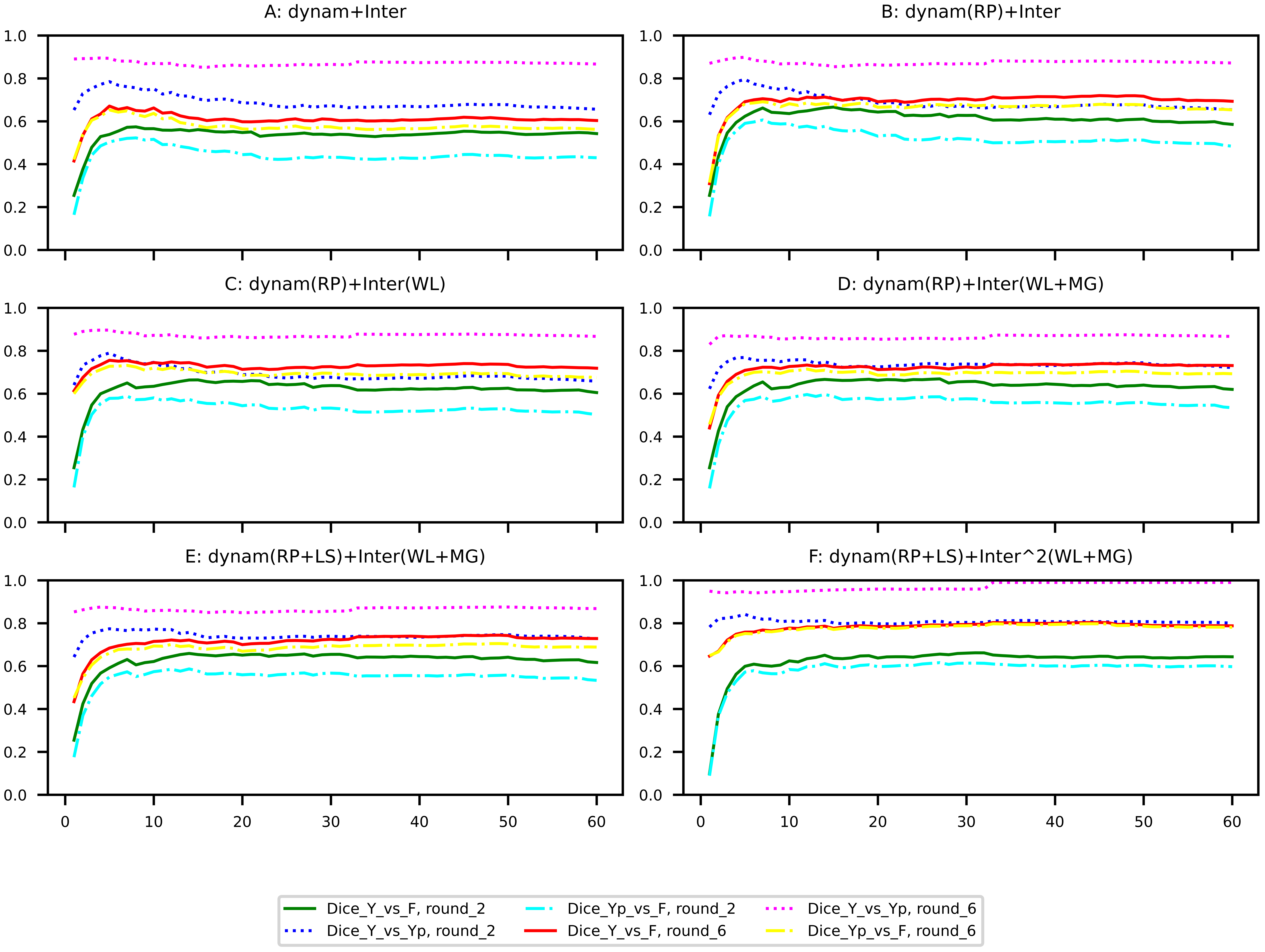}
	\caption{NCI-ISBI2013, Cum. avg. dice scores vs. number of samples in dynamic interactive training. ``Dice\_Y\_vs\_F'': $Dice(Y, F(X))$,  ``Dice\_Y\_vs\_Yp'': $Dice(Y, \Upsilon)$, ``Dice\_Yp\_vs\_F'': $Dice(\Upsilon, F(X))$; round $t$ means the $t^{\text{th}}$ round of interactive learning.}
	\label{fig:dice_triangle_brats_nci_nci}
\end{figure*}

Second, we perform dynamic interactive learning on $T_{train}$. Comparing to~\ref{sec:exp:dynamic_learning} and~\ref{sec:exp:user_interaction}, $F$ is trained with proxy masks instead of ground truth. With labeling cost equivalent to at most $\sim$38\% of the ``full efforts'', our system provided robust predictions on the evaluation set. In table~\ref{tab:dice_dynamic_interactive_nci_brats}, it is as expected that dynamic interactive learning cannot compete with the best setting in dynamic learning with ground truth labels (dynamGT(RP+LS)). However, dynamic interactive learning significantly outperforms the vanilla dynamic learning (dynamGT) (0.77/0.75 vs. 0.71/0.72). In addition, for BraTS2015, dynamic interactive learning with dynamic evaluation (dynam(RP+LS)+Inter(WL+MG)) is competitive with the best setting (warmup+MG) in table~\ref{tab:brats_interactive_compare_rl} for interactive segmentation. Note that for dynamic interactive learning we allow maximum 6 rounds of human interactions per image equivalent to at most $\sim$38\% of the workload required to draw the full target. The results demonstrated that our framework could produce state of the art results with much smaller labeling efforts! Also, the table tells similar conclusions that the replay technique (RP) and model guidance (MG) are fundamental to successful training of the system. If we allow (2+2) slices to be labeled in each round, we could reach the same level of offline benchmark with up to $\sim$76\% of the ``full efforts''. In the case of dynamic evaluation, the dice scores could reach 0.95.

\begin{table}[htb]
	\caption{DICE scores on $D_{eval}$. dynamGT means dynamic learning on $T_{train}$ \textit{with ground truth labels}; dynam+Inter means dynamic learning on $T_{train}$ with interactive segmentation (algorithm~\ref{algo:dynam_inter_learn}). By default, we allow ``users'' to draw one target slice each on coronal and sagittal planes (1+1) each round, and up to totally (6+6) (amounts to $\sim$38\% of the ``full efforts''). We also tried relaxing this constraint and allow (2+2) slices to be labeled per round ($\text{Inter}^2$).}
	\label{tab:dice_dynamic_interactive_nci_brats}
	\centering
	\begin{tabular}{lclclclcl}
		\hline
		& \multicolumn{2}{c}{Dice score: mean[min, max]} \\
		\cmidrule{2-3}
		&  NCI-ISBI2013 & BraTS2015  \\
		\hline
		\multicolumn{1}{c}{StaticTest}  & \\
		Offline       & 0.80[0.70, 0.87]   & 0.79[0.46, 0.92]  \\
		dynamGT         & 0.71[0.58, 0.80]  & 0.72[0.35, 0.90] \\
		dynamGT(RP)       & 0.82[0.72, 0.88]  & 0.79[0.46, 0.93] \\
		dynamGT(LS)    & 0.70[0.53, 0.80]  & 0.72[0.34, 0.89] \\
		dynamGT(RP+LS)   & 0.82[0.73, 0.88]   & 0.79[0.46, 0.93] \\
		\hline
		\multicolumn{1}{c}{StaticTest}  & \\
		dynam+Inter & 0.68[0.59, 0.75]  & 0.70[0.29, 0.88]   \\
		dynam(RP)+Inter & 0.75[0.63, 0.82] & 0.75[0.36, 0.90]  \\
		dynam(RP)+Inter(WL) & 0.76[0.63, 0.84]  & 0.75[0.35, 0.91]   \\
		dynam(RP)+Inter(WL+MG) & 0.77[0.62, 0.83]  & 0.75[0.43, 0.91]   \\
		dynam(RP+LS)+Inter(WL+MG) & 0.77[0.63, 0.85] & 0.75[0.39, 0.90]   \\
		dynam(RP+LS)+$\text{Inter}^{2}$(WL+MG) & 0.80[0.68, 0.88] & 0.78[0.40, 0.92] \\
		\hline
		\multicolumn{1}{c}{DynamTest}  & \\
		dynam(RP+LS)+Inter(WL+MG) & 0.84[0.73,0.91] &0.88[0.71, 0.94]   \\
		dynam(RP+LS)+$\text{Inter}^{2}$(WL+MG) & 0.95[0.92, 0.99]  & 0.95[0.90, 0.98]  \\
		\hline
	\end{tabular}	
\end{table}

It is realized that proxy masks may not be sufficiently accurate especially during earlier steps. In dynamic interactive learning, $F$ is trained towards the proxy masks instead of the unknown ground truth. An obvious concern may be that the segmenter will learn to extract some biased ``target'' induced by proxy masks, rather than the correct one. However, we found that with our framework, the segmenter implicitly learns to approximate the true labels. To see this, we track three different dice scores at each step of interactive segmentation: $Dice(Y, \Upsilon)$ and $Dice(Y, F(X))$ to measure the segmentation accuracy of proxy mask and segmenter; $Dice(\Upsilon, F(x))$ to evaluate $F$ towards its direct learning objective. Figures~\ref{fig:dice_triangle_brats_nci_brats} and~\ref{fig:dice_triangle_brats_nci_nci} shows the cumulative average of these scores in all settings of dynamic interactive learning listed in table~\ref{tab:dice_dynamic_interactive_nci_brats}. We only selected the $2^{\text{nd}}$ and $6^{\text{th}}$ rounds of the interactive segmentation for each image. We see that $D(Y, F(X))$ is consistently higher than $D(\Upsilon, F(X))$! The proxy masks of sequentially arriving images are inherently driving the segmenter to distill visual knowledge about the unknown true target. In return, the improved segmenter could also help to better guide user interactions (comparing the dashed curves of subplots C and D, in figures~\ref{fig:dice_triangle_brats_nci_brats} and~\ref{fig:dice_triangle_brats_nci_nci}.

\begin{figure*}[htb]
	\centering
	\includegraphics[width=\textwidth]{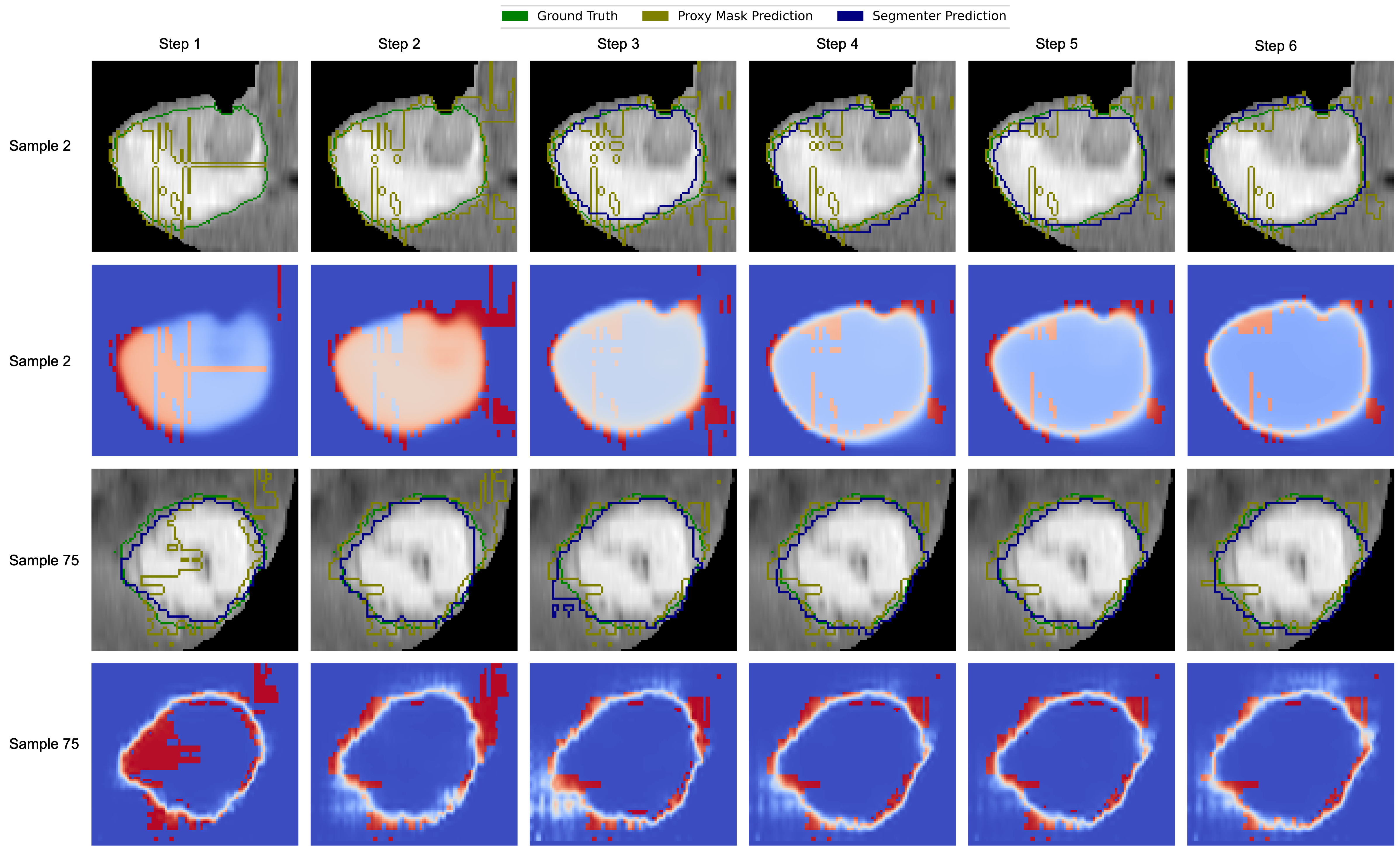} % don't need the ext-name
	\caption{\textbf{BraTS2015}. Segmentation predictions from proxy mask and segmenter during dynamic interactive learning. The first and third rows displays the predicted segmentation masks ($\Upsilon$ and $F(X)$) along with ground truth. The second and fourth rows show the heat-maps representing the residual map used for model guided annotation (equation~\ref{eq:residule_map}, algorithm~\ref{algo:propose_slice_to_label}). We choose the $2^{\text{nd}}$and $75^{\text{th}}$ images that mark early and middle stages of dynamic training; and present the results for 6 rounds of interactions. \textit{Note: we show the region cropped around the target to visualize detailed structures and shapes, the target appears smaller in raw images.}}
	\label{fig:imshowbratsinter}
\end{figure*}

\begin{figure*}[htb]
	\centering
	\includegraphics[width=\textwidth]{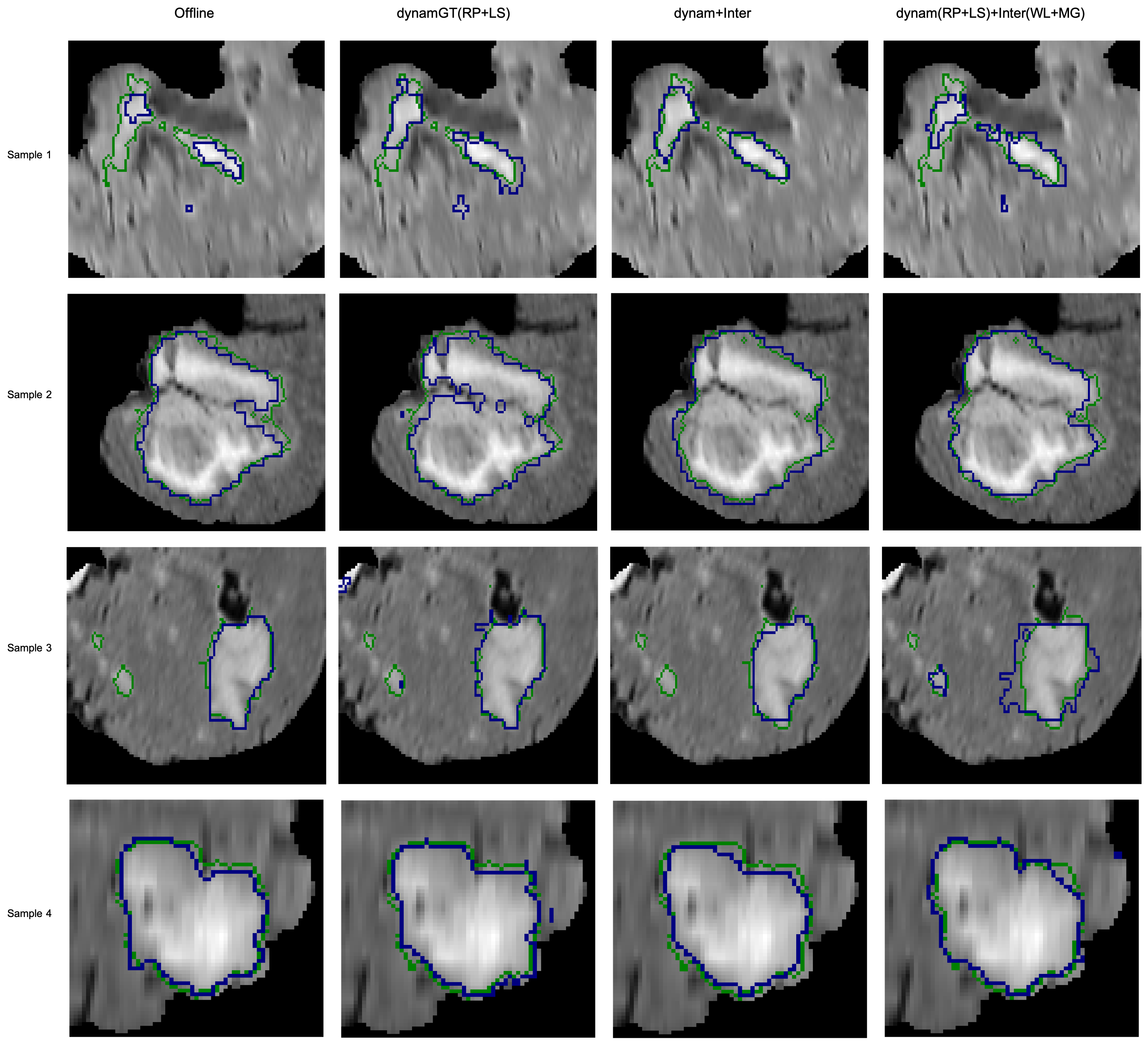} % don't need the ext-name
	\caption{\textbf{BraTS2015}. Segmentation predictions from the segmenter on four selected samples from $D_{eval}$ after dynamic interactive learning on $T_{train}$. The first column lists results from the offline benchmark. The second thru fourth columns lists settings for dynamic/dynamic-interactive learning identical to corresponding rows in table~\ref{tab:dice_dynamic_interactive_nci_brats}. \textit{Note: we show the region cropped around the target to visualize detailed structures and shapes, the target appears smaller in raw images.}}
	\label{fig:imshowbratsstdtest}
\end{figure*}

In addition the aggregated metrics, we selected representative images from $D_{eval}$ and visualize the predictions from different settings in table~\ref{tab:dice_dynamic_interactive_nci_brats}. Figure~\ref{fig:imshowbratsinter} shows the evolution of interactive segmentation for a single image during dynamic interactive training corresponding to "dynam(RP+LS)+Inter(WL+MG)" in table~\ref{tab:dice_dynamic_interactive_nci_brats}. We compare across different steps (1 thru 6) during interactive segmentation for earlier and later stages in dynamic training (the $2^{nd}$ and the $75^{th}$ samples in the sequence). Since user annotations are provided along coronal and sagittal planes, we visualize the middle slice along axial plane to avoid label leakage. From rows 1 and 3, we see that both proxy mask and segmenter prediction incrementally moved closer to the underlying ground truth boundary as more interactive steps proceeded. 

The proxy masks contain irregular structures induced by image registration. However, the segmenter supervised by proxy masks produced smooth and sufficiently accurate boundaries even at very early stage. Rows 2 and 4 show the residual map used in model guidance (algorithm~\ref{algo:propose_slice_to_label}). The high-valued areas represent locations to focus on in the next round of slice proposal. In early stages (Sample 2, Steps 1 and 2), the initial segmenter was not well trained so the map is not useful. However, starting from third step and forward, the system started pinpointing converging locations for slice selection. In later stages (Sample 75), we have a mature segmenter from the beginning of interactive learning and the similar convergence in residual maps was also observed.

Figure~\ref{fig:imshowbratsstdtest} shows the model predicted segmentation on four different evaluation samples across multiple settings in table~\ref{tab:dice_dynamic_interactive_nci_brats}. Likewise, we show the middle slice along axial plane. Column 1 shows the offline benchmark; columns 2 thru 4 correspond to dynamic learning and dynamic interactive learning. Sample 4 is a relatively easy case with smooth boundary where all four methods provided accurate approximations. For sample 2, the fourth method (dynamic interactive learning with replay and model guided annotation) provided the closest prediction comparing with the benchmark, especially that it captured subtle shapes at the right side of the target boundary. Dynamic learning with ground truth labels (dynamGT) in column 2 produced inferior predictions than both two settings of dynamic interactive learning (without ground labels). 

Samples 1 and 3 are harder cases with isolated and irregular structures. For sample 1, the offline benchmark and the third method, "dynam+Inter" gave smooth but under-segmentation, while dynamGT and the last dynamic interactive learning method generated both over/under segmentations with fewer false negative regions. The results from sample 3 show more obviously that the fourth method tend to capture isolated and smaller structures with less smoother boundaries and more false positives. These observations are likely resulted from the replay and label smoothing techniques, which yielded more robust predictions overall while imposing stricter consistency between training and evaluation set. Depending on clinical needs, if the goal is a fast approximate localization of all nodules as a first stage before finer segmentation (with user interactions), where false negatives could be tolerated on some level, our dynamic learning framework (the fourth method) could be sufficient to serve this purpose. 

In summary, from these examples we see that our dynamic interactive learning framework, even without supervised by ground truth labels, is able to generate reasonable predictions. Also, the replay technique, weighted loss and model guided annotation module could help us localize hard regions with compromises in boundary smoothness. This could be useful in cases where recall is the first concern (for example, tumor/nodule detection) and an interactive refinement could be followed. 

\subsection{Dynamic labeling in practice}\label{sec:exp:practical}

We have evaluated which level of performances could be achieved with certain limits on expert inputs. Back to one of the most practical cases: given $N$ images to be segmented, how to acquire sufficiently accurate segmentations for \textit{all of them} as fast as possible (i.e. solving equation~\ref{eq:inter_learning_opt_21} for all $N$ images). All of previous evaluations, including other state-of-the-arts~\cite{ref_23}, focused on average metrics. Now we need to ensure that the minimal dice scores beyond a certain threshold $\rho$. We set $\rho=0.85$ in the experiment.

The most tedious way would be to delineate the entire 3D target slice by slice for all images; which marks the upper bound of user time $C_{\text{max}}=\sum_{i=1}^N C(Y_i)$. A ``semi-tedious'' approach is to fully label part of the images, train a segmenter and generate predicted masks for the rest. This won't work in practice since we never know when to stop labeling to guarantee the required accuracy for all remaining images. In fact, the offline benchmark in table~\ref{tab:dice_dynamic_brats_nci} showed the minimum dice scores of 0.70/0.46 for the test set of two data sets, even with the majority of samples labeled for training! We will see how the dynamic interactive learning framework solves the problem. 

\begin{figure*}[htb]
	\centering
	\includegraphics[width=\textwidth]{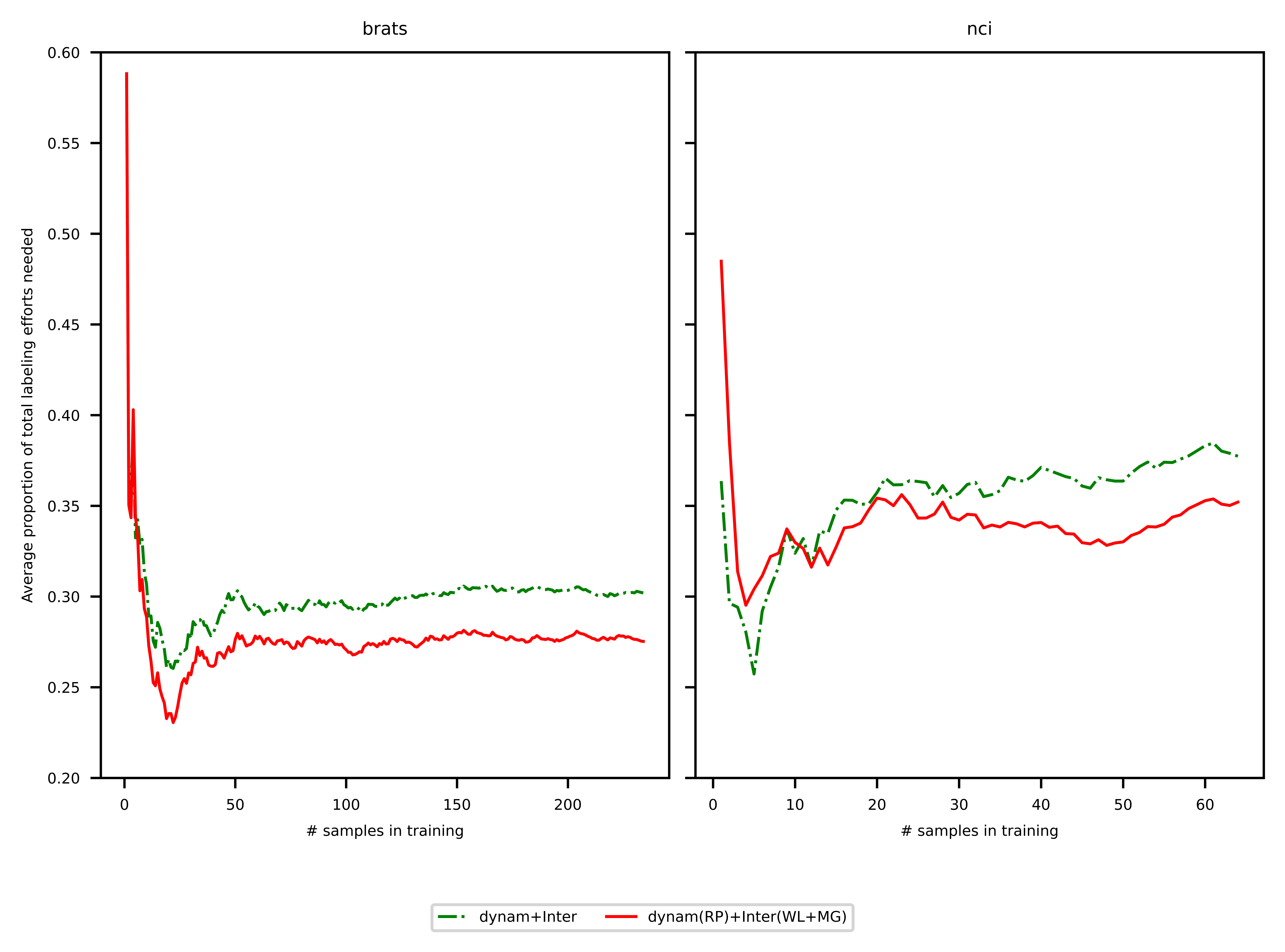} % don't need the ext-name
	\caption{Cumulative average proportion of user labors needed vs. number of samples in dynamic interactive training.}
	\label{fig:dice_reachmode_brats_nci}
\end{figure*}

Table~\ref{tab:dice_reachmode_brats_nci} lists the user labor needed to have accurate segmentations for all images in training set. Images enter the dynamic interactive learning framework sequentially, and interactive segmentation will keep running until we have a proxy mask whose dice $>\rho$. The setting "dynam+Inter" serves as the baseline where users randomly selected slices to label without model guidance. We see that model guided annotation is critical in reducing user labor. Figure~\ref{fig:dice_reachmode_brats_nci} shows more straightforwardly the proportion of required user labors to the ``full efforts''. We see that with replay and model guidance, the dynamic interactive learning framework effectively reduces the total user labor. In addition, as the segmenter learned from more samples in the sequence, the reduction of user labor become more significant (the gap between red and green curves). Moreover, a rapid drop of the red curve tells us the underlying segmenter trained in our way is able to capture generic patterns of the segmentation task pretty fast, which is consistent to the observations in figures~\ref{fig:dice_triangle_brats_nci_brats} and~\ref{fig:dice_triangle_brats_nci_nci}.

\begin{table*}[htb]
	\caption{Minimum user labor needed in interactive segmentation s.t. the dice score for each of $T_{train}$ is beyond $\rho$. The user labor is measured by number of slices labeled in coronal and sagittal planes (C+S). Cumulative metrics is computed as 20\%, 40\%, 60\%, 80\% and 100\% of $T_train$ enter the system. "Cum. full labeling (C/S)" is the number of slices on coronal and sagittal planes that constitutes the full target.}
	\label{tab:dice_reachmode_brats_nci}
	\centering
	\begin{tabular}{lclclclclclc|}
		\hline
		& \multicolumn{5}{c}{Cum. user labors needed ($\#$ C + S slices labeled)} \\
		\cmidrule{2-6}
		$\#$ Prop. samples in training   &20\%  &40\%  &60\%  &80\% &100\%  \\
		\hline
		\multicolumn{1}{c}{BraTS2015}  &   &  &  &  & \\
		$\#$ Cum. full labeling (C/S) &1066/2008 &2149/4048 &3232/6101 &4318/8134 &5396/10168  \\
		dynam+Inter    &229+229   &461+461  &704+704 &949+949  &1175+1175  \\
		dynam(RP)+Inter(WL+MG) &207+207  &428+428  &645+645   &864+864  &1071+1071    \\
		\hline
		\multicolumn{1}{c}{NCI-ISBI2013}  &   \\
		$\#$ Cum. full labeling (C/S) &264/533 &464/1121 &672/1716 &873/2268 &1102/2852  \\
		dynam+Inter    &63+63  &144+144   &217+217    &289+289  &373+373  \\
		dynam(RP)+Inter(WL+MG) &63+63  &136+136   &202+202    &262+262  &348+348  \\
		\hline
	\end{tabular}	
\end{table*}

\section{Conclusion}
This work presents a novel dynamic interactive learning framework to address practical challenges of 3D medical image segmentation. The framework effectively deals with sequentially arriving images and achieved sufficiently accurate predictions with substantially reduced user annotation cost. To the best of our knowledge, this paper presents the first efforts to combine dynamic learning with interactive segmentation, and to thoroughly discuss each individual components for optimal learning strategies. We have shown that the dynamic learning module, with the proposed replay mechanism, could compete with the offline benchmark. Also, with model guided annotation, interactive segmentation module could quickly produce competitive results (comparing with state-of-the-arts of similar settings) even with an un-trained segmenter from the beginning. Also, without knowing the ground truth, dynamic interactive learning could generate robust performances that approach the offline benchmark with incremental sparse annotations up to $\sim$38\% of the total efforts needed for labeling the full target. Our framework is also flexible in supporting multiple back-bone segmenter architectures as well as segmentation tasks in various scenarios (with/without user interventions, cold-start cases, labeling streaming data, etc.). Due to its adaptive and interactive nature, it could be deployed into the hospital firewall that provide higher level of data security and engineering efficiency. Further improvements on the framework could also be explored, as part of our future directions, including supporting other types of annotations (clicks/free-form stripes) for user intervention, and making the model guidance scheme (algorithm~\ref{algo:propose_slice_to_label}) as a differentiable procedure, etc. We believe our proposed framework could provide insights that benefit both engineers and physicians in research and clinical applications.

\paragraph{Acknowledgment}
This work was supported in part by the Key-Area Research and Development Program of Guangdong Province grant 2021B0101420005,the Shenzhen Natural Science Fund (the Stable Support Plan Program20220810144949003), the Key Technology Development Program of Shenzhen grant JSGG20210713091811036, the Shenzhen Key Laboratory Foundation grant ZDSYS20200811143757022, the SZU Top Ranking Project grant 86000000210, and the National Natural Science Foundation of China grant 62301326.

\bibliographystyle{unsrt}  
\bibliography{references}  %%% Remove comment to use the external .bib file (using bibtex).


\end{document}